\documentclass[final,5p,times,twocolumn,authoryear]{elsarticle}
\usepackage{fontawesome}

\usepackage{amssymb,amsmath}
\usepackage{lipsum}
\usepackage{booktabs}
\usepackage{diagbox}
\usepackage{hyperref}

\usepackage{comment}

\usepackage{color}

\journal{Nuclear Physics B}

\begin{document}

\begin{frontmatter}



\title{Comparing Generative Models with the New Physics Learning Machine}

\affiliation[label1]{organization={Department of Physics, University of Genova},
            addressline={Via Dodecaneso 33}, 
            city={Genova},
            postcode={I-16146}, 
            state={},
            country={Italy}}
            
\affiliation[label2]{organization={INFN, Sezione di Genova},
            addressline={Via Dodecaneso 33}, 
            city={Genova},
            postcode={I-16146}, 
            state={},
            country={Italy}}

\affiliation[label3]{organization={MaLGa-DIBRIS, University of Genova},
            addressline={Via Dodecaneso 35}, 
            city={Genova},
            postcode={I-16146}, 
            state={},
            country={Italy}}

\author[label1,label2]{Samuele Grossi}

\author[label2,label3]{Marco Letizia}

\author[label1,label2]{Riccardo Torre}

\begin{abstract}
The rise of generative models for scientific research calls for the development of new methods to evaluate their fidelity. A natural framework for addressing this problem is two-sample hypothesis testing, namely the task of determining whether two data sets are drawn from the same distribution. In large-scale and high-dimensional regimes, machine learning offers a set of tools to push beyond the limitations of standard statistical techniques. In this work, we put this claim to the test by comparing a recent proposal from the high-energy physics literature, the New Physics Learning Machine, to perform a classification-based two-sample test against a number of alternative approaches, following the framework presented in \cite{Grossi:2024axb}. We highlight the efficiency tradeoffs of the method and the computational costs that come from adopting learning-based approaches. Finally, we discuss the advantages of the different methods for different use cases.
\end{abstract}




\end{frontmatter}




\section{Introduction}
\label{introduction}

The growing importance of generative models to produce high-dimensional synthetic data cannot be understated, in both scientific and industrial domains. In the context of precision sciences, such as High Energy Physics (HEP), they offer a promising route to accelerate simulations traditionally handled by high-fidelity Monte Carlo methods (e.g. \cite{Butter:2022rso,Krause:2024avx}). However, ensuring their reliability requires validation techniques that can match the precision of existing simulation pipelines.

The problem of testing a candidate generative model against a reference data-generating method can naturally be framed as a goodness-of-fit (GoF) test, which aims at establishing whether a statistical model adequately describes a set of data. This approach is enabled by the assumption that the reference model can be sampled at will, which allows full characterization of the null hypothesis of the statistical test through empirical samples rather than analytical expressions. In this data-driven setup, the GoF test can then be practically implemented as a two-sample test (2ST).
Several proposals to address the task of evaluating generative models in science have been discussed in the literature, and the question of how to assess the advantages and limitations of different approaches has become increasingly relevant (see~\cite{Das:2023ktd,Kansal:2022spb,Grossi:2024axb,Cappelli:nplm2024}). Standard statistical tests often struggle to fully capture the complexity of modern scientific data, especially when dealing with large sample sizes or high-dimensional spaces. From this perspective, developing testing strategies based on machine learning represents a promising direction. The New Physics Learning Machine (NPLM) (\cite{DAgnolo:2018cun, DAgnolo:2019vbw, Letizia:2022xbe, Grosso:2023scl}) is a compelling classifier-based methodology designed to perform a data-driven likelihood-ratio test, primarily for HEP data.
A potential drawback of learning-based methods is the introduction of a training step that inevitably impacts the efficiency of the test in terms of computational time. It then becomes relevant to assess the tradeoff between sensitivity and efficiency and determine whether non-learning methodologies retain an advantage, especially when generative models have typically not yet reached a high degree of fidelity by the standards of precision sciences. Moreover, machine learning methods generally require a model selection phase that could further affect their overall efficiency.

In this work, we aim to deploy the framework developed in~\cite{Grossi:2024axb} to assess the performance of the NPLM method against several other evaluation metrics previously proposed in the literature. While NPLM has been primarily tested on relatively low-dimensional problems (around ten features), as is common in many HEP applications, we also evaluate its performance in more challenging regimes with dimensionalities up to $d=100$. We adopt here balanced samples from the reference and generative distributions to enable a clean and controlled comparison across evaluation methods. Finally, we highlight some details about the model selection pipeline that are missing from the previous literature and discuss its overall impact on the efficiency of the test compared to the alternatives.

Code and full results are available on GitHub in~\cite{NPLM_Gauss_results, NPLM_JetNet_results,NPLM_Parameters_Tuning}.

\section{Comparing generators through two-sample tests}\label{sec:method-summary}

In this section, we briefly summarize the methodology for comparing generators via two-sample testing, following the framework introduced in~\cite{Grossi:2024axb}.

Let $\mathcal{G}_{p}$ denote a reference generator producing samples from a known probability density function (pdf) $p$. To test whether another generator $\mathcal{G}'_{q}$ produces samples consistent with $\mathcal{G}_{p}$, we compare two datasets $\mathcal{X}=\{x_i\}_{i=1}^n$ and $\mathcal{Y}=\{y_j\}_{j=1}^m$, produced by the two generators, using a two-sample test. The null hypothesis $H_0$ states that both datasets are generated by the reference generator $\mathcal{G}_{p}$.

A test statistic $t$ for a two-sample test is defined as a scalar-valued function that maps two samples of sizes $n$ and $m$ in $d$ dimensions into a real number:
\begin{equation}\label{eq:test_stat}
t : \mathbb{R}^{n \times d} \times \mathbb{R}^{m \times d} \to \mathbb{R}.
\end{equation}
Its distribution under $H_0$ is estimated empirically by computing $t$ over multiple pairs of samples independently generated by $\mathcal{G}_{p}$. This results in a set of values of $t_{0}$ from which we construct the empirical pdf $f(t_0)$ and cumulative distribution function (cdf) $F(t_0)$. These define the significance threshold $t_0^{\alpha}$ corresponding to a fixed significance $1-\alpha$:
\begin{equation}
    \alpha= \int_{t_{0}^{\alpha}}^{\infty}dF(t_{0}) = \int_{t_{0}^{\alpha}}^{\infty}f(t_{0})dt_{0},
\end{equation}
which we estimate empirically as:
\begin{equation}
\begin{split}
1 - \alpha = F(t_0^\alpha) &\approx 1 - \frac{\# \text{ of tests with } t_0 \geq t_0^\alpha}{\text{total }\# \text{ of tests}}\\
&= \frac{\# \text{ of tests with } t_0 < t_0^\alpha}{\text{total }\# \text{ of tests}}.
\end{split}
\end{equation}
In this work, we consider $\alpha = 0.05$ and $0.01$.

To assess the sensitivity of a test statistic to deviations from $p$, we define alternative generators $\mathcal{G}'_{q_\epsilon}$, obtained by deforming the pdf $p$ through a scalar parameter $\epsilon$, namely
$$
q_\epsilon\xrightarrow{\epsilon\rightarrow0}p.$$
These $\epsilon$-deformations are designed to systematically test the sensitivity of each test statistic, and are listed in~\ref{app:deformations}. The corresponding alternative hypothesis $H_1$ states that $\mathcal{X} \sim p^n$ and $\mathcal{Y} \sim q^m_\epsilon$ are generated by different generators ($\epsilon\neq 0$).

The critical deformation $\epsilon_\alpha$ is defined as the smallest $\epsilon\geq 0$ that can be ``detected" (leading to rejection of the null-hypothesis) by a given test with a significance level $1-\alpha$. This corresponds to the smallest $\epsilon$-deformation for which the test statistic exceeds the threshold $t_0^\alpha$ and can be formulated as the following optimization problem:
\begin{equation}\label{eq:opt_problem}
\epsilon_\alpha = \arg\min_{\epsilon} |t(\epsilon) - t_0^\alpha|.
\end{equation}
We solve this numerically using a simple bisection method. At each step, the test is repeated 100 times to estimate the mean $\mu_{t(\epsilon)}$ and standard deviation $\sigma_{t(\epsilon)}$ of the test statistic. The interval for $\epsilon$ is iteratively refined by comparing $\mu_{t(\epsilon)} \pm \sigma_{t(\epsilon)}$ to $t_0^\alpha$, until convergence is reached within a fixed tolerance of $10^{-2}$. The central value defines $\epsilon_\alpha$, with bounds $\epsilon_{\alpha\text{-low}}$ and $\epsilon_{\alpha\text{-up}}$ determined by where the uncertainty bands meet the threshold. In formulae, this is expressed as:
\begin{equation}
\begin{split}
&\epsilon_{\alpha} = \arg\min_{\epsilon} |\mu_{t(\epsilon)} - t_0^\alpha|,\\
&\epsilon_{\alpha\text{-low}} = \arg\min_{\epsilon} |(\mu_{t(\epsilon)}+ \sigma_{t(\epsilon)}) - t_0^\alpha|,\\
&\epsilon_{\alpha\text{-up}} = \arg\min_{\epsilon} |(\mu_{t(\epsilon)}- \sigma_{t(\epsilon)}) - t_0^\alpha |.
\end{split}
\end{equation}

This framework applies to non-parametric test statistics, which do not require explicit knowledge of $p$ or $q_\epsilon$. However, we also include the log-likelihood ratio (LLR) test, which requires both pdfs to be known. Its test statistic is defined as:
\begin{equation}
t(\epsilon) = -2 \sum_{y \in \mathcal{Y}} \log \frac{p(y)}{q_\epsilon(y)}.
\end{equation}
This, by construction, only depends on the samples $\mathcal{Y}$ which are generated by $\mathcal{G}_{p}$ in the case of the null hypothesis $H_0$, and by $\mathcal{G}'_{q_\epsilon}$ in the case of the alternative hypothesis $H_1$. Moreover, since this test statistic depends explicitly on $q_\epsilon$, the null distribution $f(t_0)$ becomes $\epsilon$-dependent. Accordingly, the optimization problem becomes:
\begin{equation}
\epsilon_\alpha = \arg\min_{\epsilon} |t(\epsilon) - t_0^\alpha(\epsilon)|,
\end{equation}
where $t_0^\alpha(\epsilon)$ must be computed for each value of $\epsilon$. When applicable, the LLR provides the most powerful test according to the Neyman–Pearson lemma (\cite{Neyman:1933wgr}).

Finally, in many practical applications the generators $\mathcal{G}$ may not provide a closed analytical form for the pdf. In such cases, only a finite number of samples is available, and the LLR test is not applicable. The distribution $f(t_0)$ must then be estimated empirically by using a bootstrap approach (namely, sampling with replacement) to mimic the behavior of $\mathcal{G}_{p}$ and $\mathcal{G}'_{q_\epsilon}$.

\section{The NPLM method}
NPLM is a machine learning-based, signal-agnostic hypothesis testing approach designed on the basis of the maximum likelihood-ratio test as formulated by~\cite{Neyman:1933wgr}. Originally developed for the discovery of new physics in high-energy collider experiments such as the LHC (\cite{DAgnolo:2018cun}), we consider it here as a general-purpose testing methodology for comparing data generators.

At its core, the NPLM method leverages the ability of classifiers to estimate the ratio of data-generating pdfs (see e.g.~\cite{hastie2009elements}). Adopting the notation from the previous section, 
a classifier is trained to approximate the following function\footnote{Because the original NPLM framework was developed for LHC analyses, these densities are typically normalized to different expected event counts. This distinction is not relevant for our discussion.}
\begin{equation}
f_{\hat{w}}(z) \approx \log \frac{q(z)}{p(z)},
\end{equation}
where $\hat{w}$ are the model parameters obtained at the end of training. 
The model is then evaluated in-sample on the full dataset using the metric
\begin{equation}\label{eq:nplm_statistic}
t_{\rm NPLM}(\mathcal{X}, \mathcal{Y}) = -2\left[\frac{m}{n} \sum_{z \in \mathcal{X}} \left(e^{f_{\hat{w}}(z)} - 1\right) - \sum_{z \in \mathcal{Y}} f_{\hat{w}}(z)\right],
\end{equation}
which represents a Monte Carlo-based formulation of the extended likelihood ratio (see~\cite{barlow1990extended,DAgnolo:2018cun,Letizia:2022xbe} and~\ref{app:ext_lr}). The NPLM method operates as a two-sample test, taking $\mathcal{X}$ and $\mathcal{Y}$ as inputs and returning a scalar value in accordance with Eq.~\eqref{eq:test_stat}. It then integrates naturally within the framework introduced in~\cite{Grossi:2024axb} to assess its performance against other approaches. Due to its training efficiency, we employ the implementation presented in~\cite{Letizia:2022xbe}, in which the learning model is based on kernel methods and the classifier spans a parameterized function space \( \mathcal{F} = \{f_w\} \), defined as a weighted sum of Gaussian kernels:
\begin{equation}\label{kernel_methods}
f_w(z) = \sum_{i=1}^{n+m} w_i k_\sigma(z, z_i), \quad k_\sigma(z, z') = \exp\left(-\frac{\|z - z'\|^2}{2\sigma^2}\right),
\end{equation}
where the kernel width \( \sigma \) is treated as a hyperparameter. The loss function is the weighted binary cross-entropy loss:
\begin{equation}\label{eq:logistic}
\ell(c, f_w(z)) = (1 - c)\frac{m}{n} \log\left(1 + e^{f_w(z)}\right) + c \log\left(1 + e^{-f_w(z)}\right),
\end{equation}
where $\mathcal{Z} = \{z_i\}_{i=1}^{n+m} = \{x_1,\dots,x_n,y_1,\dots,y_m\}$ and class labels are defined as $c = 0$ for $z \in \mathcal{X}$ and $c = 1$ for $z \in \mathcal{Y}$. The model is trained to minimize the empirical risk:
\begin{equation}\label{reg_ERM}
L(f_w) = \frac{1}{n + m} \sum_{i=1}^{n + m} \ell(c_i, f_w(z_i)) + \lambda R(f_w),
\end{equation}
where $R(f_w)$ is the analogue of $L^2$ regularization in the context of kernel methods (\cite{smola1998learning}).

Despite its effectiveness, this approach can have high computational costs when the sample size is large. To mitigate this issue, the authors of \cite{Letizia:2022xbe} rely on Falkon (\cite{meanti2020kernel}), a modern solver for large-scale kernel methods. Falkon replaces Eq.~\eqref{kernel_methods} with:
\begin{equation}\label{nystrom_km}
f_w(z) = \sum_{i=1}^M w_i k_\sigma(z, \tilde{z}_i),
\end{equation}
where \( \{\tilde{z}_1, \dots, \tilde{z}_M\} \), known as Nyström centers, are sampled uniformly at random from the full dataset. The parameter \( M \) is a tunable hyperparameter.

\subsection{Hyperparameter Tuning}\label{subsec:NPLMtuning}
The first step in the NPLM methodology is hyperparameter tuning. 
The kernel-based implementation of NPLM involves three primary hyperparameters: the kernel width $\sigma$, the regularization parameter $\lambda$, and the number of centers $M$. These are tuned exclusively using reference data. Following~\cite{Letizia:2022xbe}, the selection criteria are:
\begin{itemize}
    \item The Gaussian kernel width $\sigma$ is set to the 90th percentile of the pairwise distances among reference-distributed data points. Heuristics of this type are commonly used in kernel methods (\cite{gretton2012kernel}).

    \item The regularization parameter $\lambda$ is chosen to be as small as possible, subject to computational constraints and while ensuring stable training dynamics (\cite{rudi2016more}).
    
    \item The number of Nyström centers $M$ should be at least of order $\sqrt{n+m}$ (\cite{rudi2016more}). Larger values of $M$ improve performance but increase computational costs, both in terms of training time and memory. For small $M$, the test statistic increases with $M$ until it reaches a plateau. We therefore require that the average value of the test statistic — computed over a small number of tests on reference data (i.e., under $H_0$) — be approximately stable as a function of $M$.
\end{itemize}

We will show the results of this pipeline on our data in the next section.

\section{Numerical analysis}   

In this section, we present the results of the analysis based on~\cite{Grossi:2024axb}, and summarized in Section~\ref{sec:method-summary}, when applied to the NPLM method. We start by summarizing the properties of the datasets we consider (see~\cite{Grossi:2024axb} for a in-depth description) and we then detail the hyperparameter tuning of the NPLM method. This is an important point of deviation with respect to standard testing methodologies. 

\subsection{Data}

\begin{description}
\item[Mixtures of Gaussians (MoG)] We consider mixtures of $q$ multivariate Gaussians ($q$ components) in $d$ dimensions, each with diagonal covariance matrices. This framework enables the study of probability density functions with multiple local maxima, which manifest as multiple peaks in the marginal distributions. For our analysis, we examine three MoG configurations: $q = 3$ components in $d = 5$ dimensions, $q = 5$ components in $d = 20$ dimensions, and $q = 10$ components in $d = 100$ dimensions.
\item[Correlated Gaussians] These are correlated $d$-dimensional unimodal Gaussian distributions. In our analysis, we consider $d=5,20,100$.
\item[JetNet] To explore a scenario relevant for HEP, we also consider a dataset of simulated gluon jets from the JetNet dataset. We examine two complementary data representations: a particle-level dataset (90 dimensions) that includes features of individual particles within each jet, and a jet-level dataset (3 dimensions) that captures only high-level jet characteristics.
\end{description}

\subsection{Hyperparameter tuning}

The choice of hyperparameters determines the complexity of the learning model. In this work, we perform model selection aiming for a reasonable trade-off between model complexity and computational efficiency, and we do not target the most complex model that can be computed given available hardware resources.
\begin{figure}[t!]
    \centering
    \includegraphics[width=0.75\linewidth]{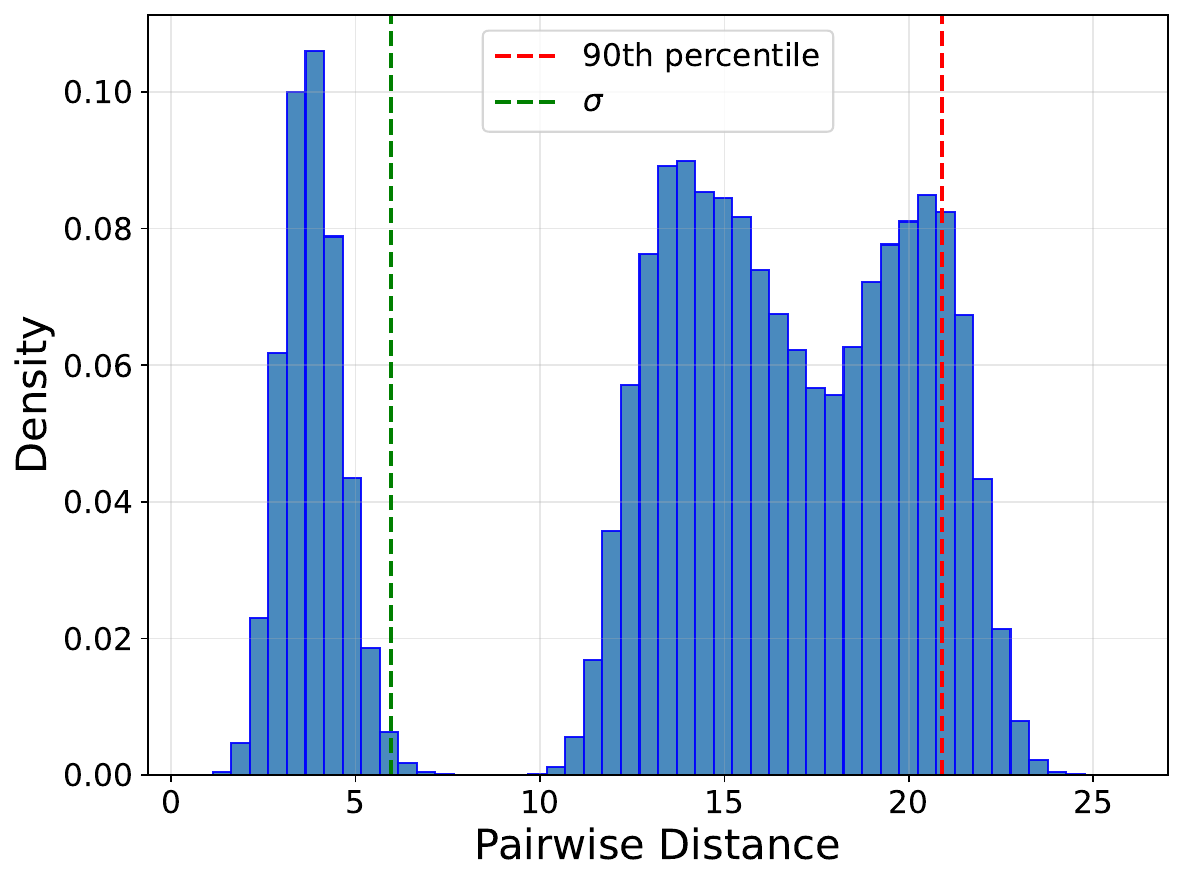}
    \caption{Example of a multimodal distribution of the pairwise distance from the MoG model in 20 dimensions.}
    \label{fig::tuning_sigma}
\end{figure}
\begin{table}[t!]
\small
\resizebox{0.49\textwidth}{!}{
\begin{tabular}{l|ccc}
    \toprule
    \multicolumn{1}{c}{} & \multicolumn{3}{c}{Mixture of Gaussians} \\
    \bottomrule
    {\diagbox{$n$}{$d$}} & 5 & 20 & 100 \\
    \toprule
    $10K$   & $(3.88,10000,10^{-8})$ & $(5.98,13500,10^{-7})$ & $(10.31,16000,10^{-5})$ \\
    $20K$   & $(3.88,7000,10^{-8})$  & $(5.98,12500,10^{-7})$ & $(10.31,16000,10^{-5})$ \\
    $50K$   & $(3.88,7000,10^{-8})$  & $(5.98,12500,10^{-7})$ & $(10.31,16000,10^{-5})$ \\
    $100K$  & $(3.88,5000,10^{-8})$  & $(5.98,11000,10^{-6})$ & $(10.31,11000,10^{-5})$ \\
    \bottomrule
\end{tabular}
}
\caption{Values of the hyperparameters $(\sigma,M,\lambda)$ for the MoG distributions for the different dimensionalities $d$ and sample sizes $n$.}
\label{tab::distributions_parameters_MoG}
\end{table}

We start by considering the kernel bandwidth $\sigma$. As discussed in the NPLM literature, the distribution of pairwise Euclidean distances between reference-distributed points is considered as a method to estimate the most relevant scales in the problem, and it was observed that selecting $\sigma$ as the 90th percentile of this distribution allows one to obtain a statistical test that has reasonably homogeneous sensitivity over a wide range of potential anomalies. However, this distribution can in general be multimodal, signaling that the reference distribution is characterized by more than one scale. We then select this particular hyperparameter relying on a straightforward modification of the methodology outlined in the previous section. If the distribution of pairwise distances has multiple peaks, as in the case of data from the MoG models, we focus on the first peak (the one at the smallest value of pairwise distance). In particular, we take a value lying on the right tail of the peak.
We do not fine-tune this hyperparameter further, as its exact value is not expected to meaningfully affect the sensitivity of the test. 
This approach is illustrated in Figure~\ref{fig::tuning_sigma} for the MoG in 20 dimensions.
\footnote{See also~\cite{Grosso:2024wjt} for a more refined, but with higher computational costs, approach to this problem.}

\begin{figure}[t!]
    \centering
    \includegraphics[width=\linewidth]{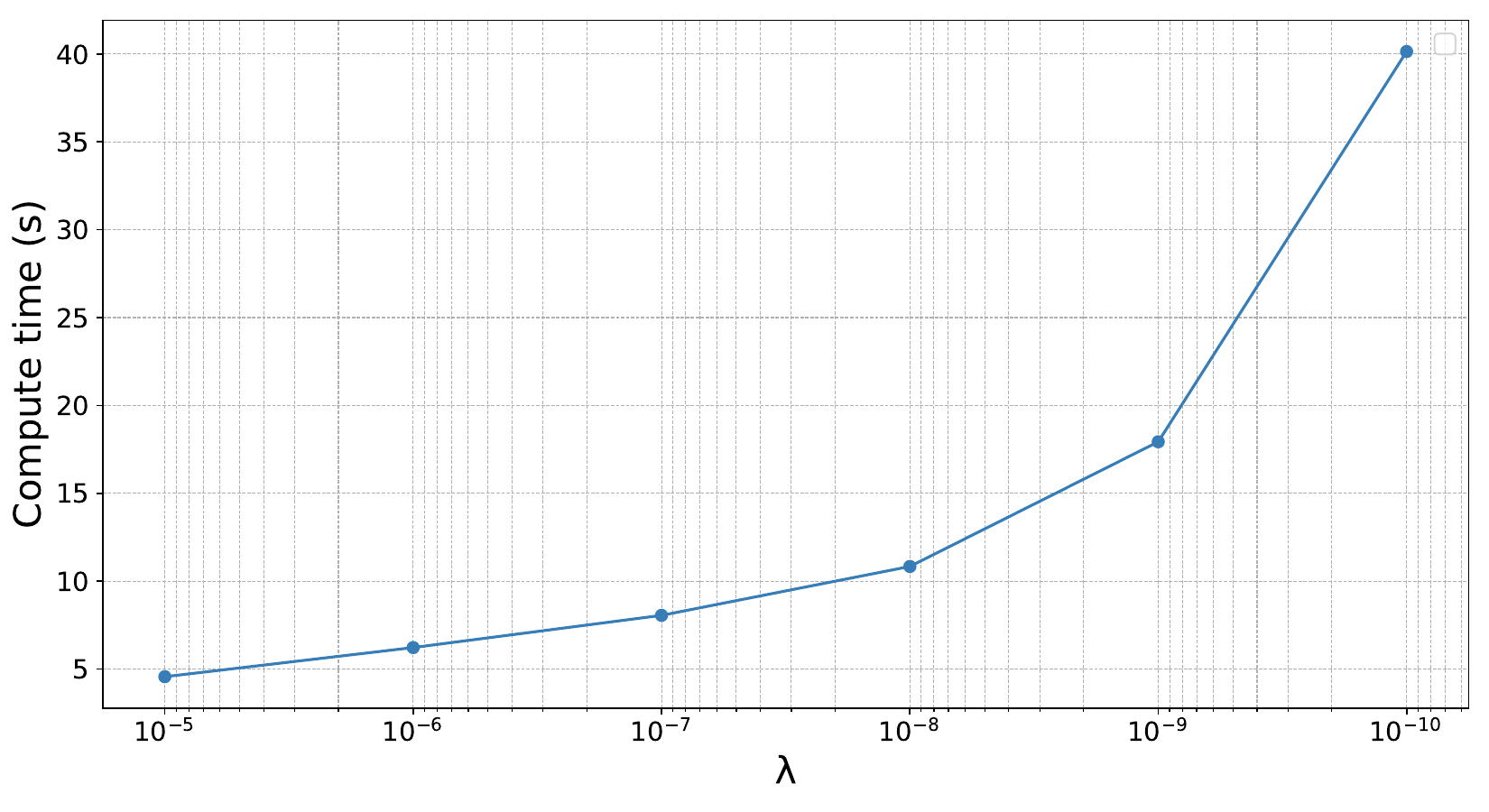}
    \caption{Mean compute time over 20 tests as a function of $\lambda$ for the MoG model with $d=20$ and $n=50K$.}
    \label{fig::Computing_time_vs_lambda_MoG_20D_50K}
\end{figure}

\begin{figure*}[t!]
    \centering
    {
        \includegraphics[width=0.48\linewidth]{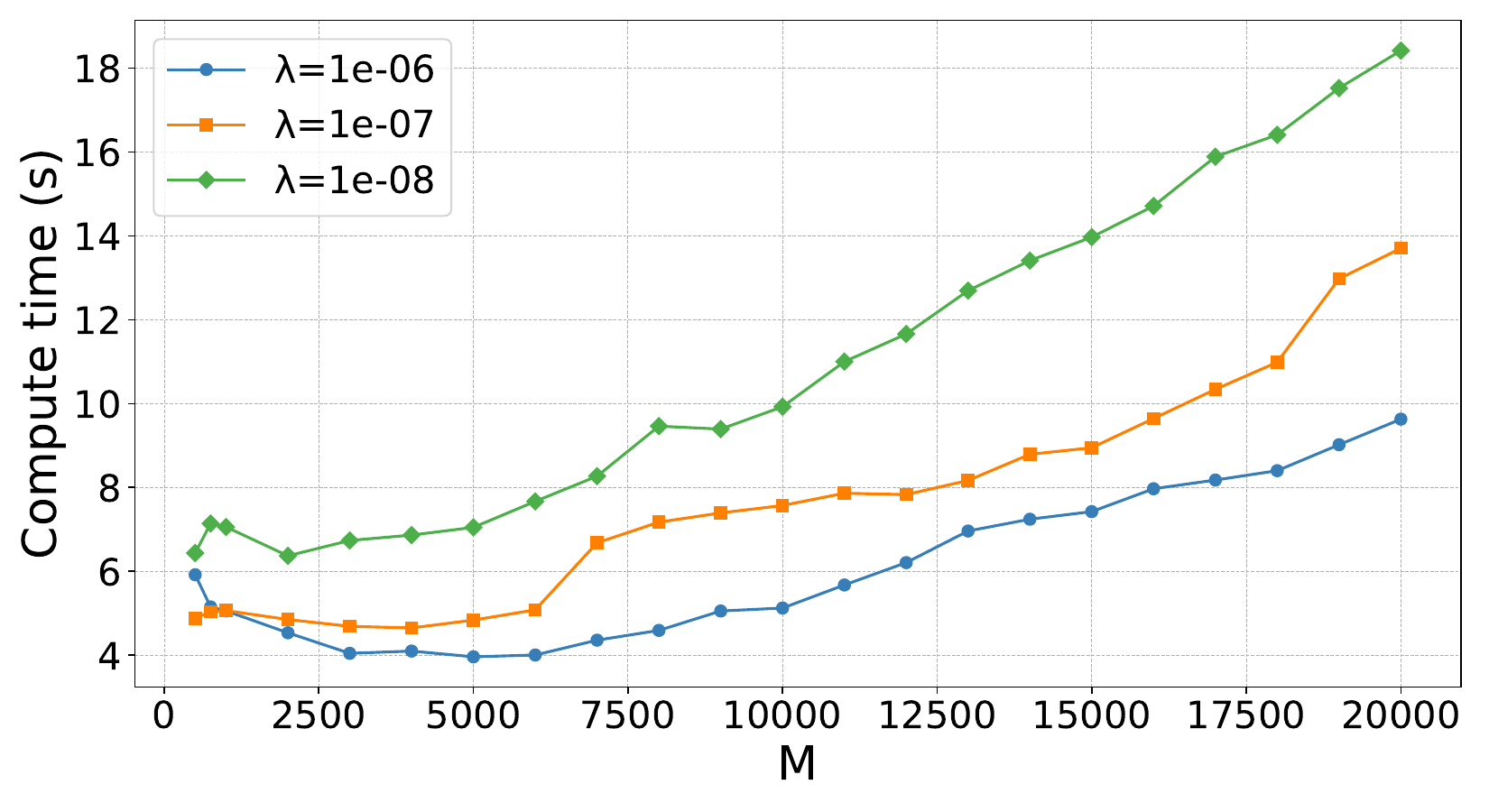}
    }
    \centering
    {
        \includegraphics[width=0.48\linewidth]{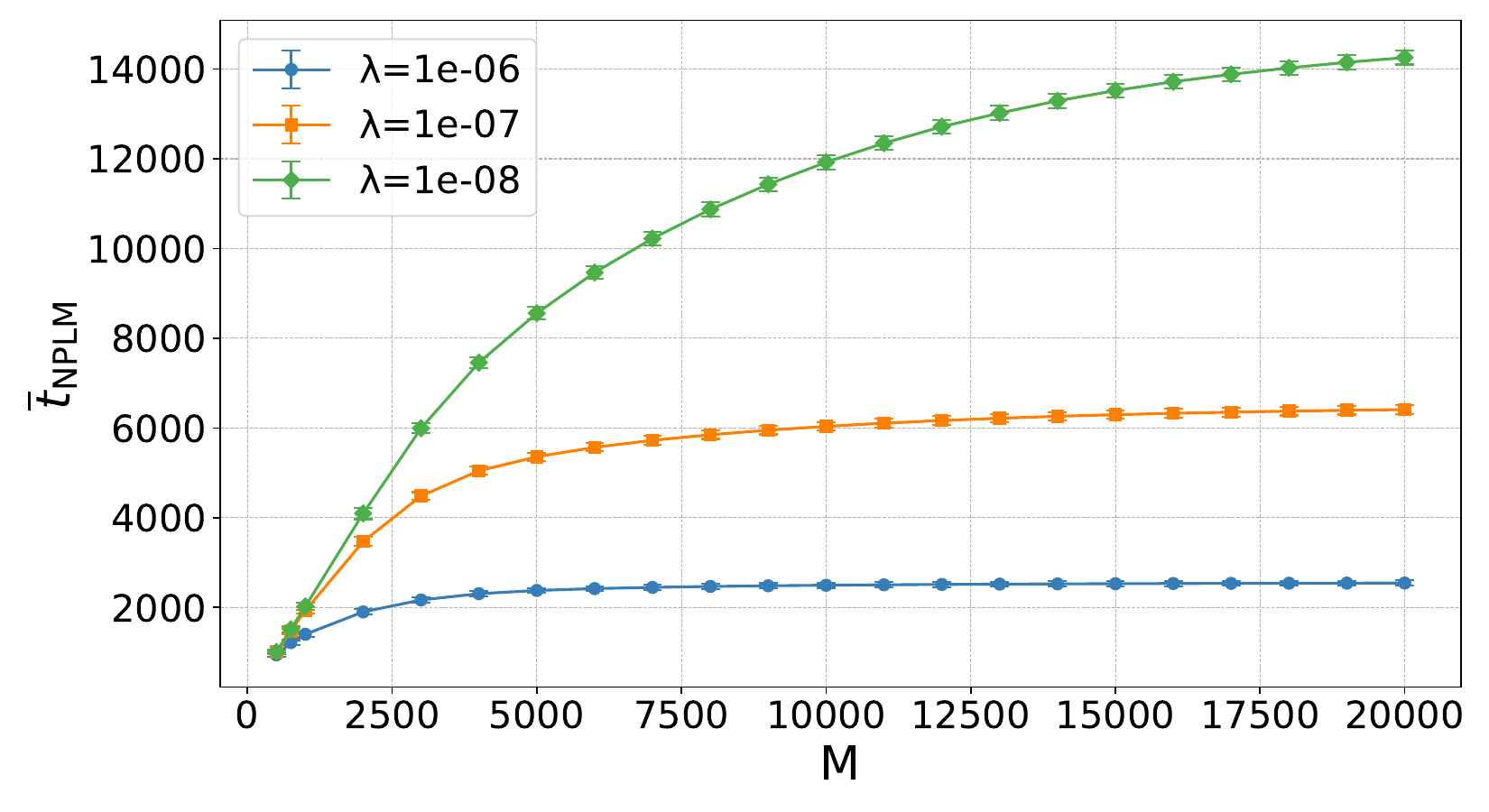}
    }
    \caption{Compute time and mean test statistic as functions of $M$ at varying $\lambda$ for the MoG model with $d=20$ and $n=50K$.}
    \label{fig::comp_test_vs_M_MoG_20D_50K}
\end{figure*}

To explore the space of $M$ and $\lambda$ values, we compute the average NPLM test statistic over 20 randomly selected pairs of samples drawn from the reference distribution. We first estimate the impact of $\lambda$ on the training time. The parameter tuning for this analysis has been performed selecting $M=1000,2000,3000,5000$ for $n=10K,20K,50K,100K$. 
We report in Figure~\ref{fig::Computing_time_vs_lambda_MoG_20D_50K} an example of the computing time as a function of $\lambda$ for the Mixture of Gaussians (MoG) model with $d=20$ and $n=50K$, an intermediate representative case.
\begin{table}[t!]
\small
\resizebox{0.49\textwidth}{!}{
\begin{tabular}{l|ccc}
	\toprule
	\multicolumn{1}{c}{} & \multicolumn{3}{c}{Correlated Gaussians} \\
    \bottomrule
	{\diagbox{$n$}{$d$}} & 5 & 20 & 100 \\
	\toprule
	$10K$ & $(9.8,10000,10^{-8})$ & $(21.8,13500,10^{-7})$ & $(52.6, 16000,10^{-5})$ \\
	$20K$ & $(9.8,7000,10^{-8})$ & $(21.8,12500,10^{-7})$ & $(52.6, 16000,10^{-5})$ \\
	$50K$ & $(9.8,7000,10^{-8})$ & $(21.8,12000,10^{-7})$ & $(52.6, 16000,10^{-5})$ \\
	$100K$ & $(9.8,5000,10^{-8})$ & $(21.8,11000,10^{-6})$ & $(52.6, 14000,10^{-5})$ \\
	\bottomrule
\end{tabular}
}
\caption{Values of the hyperparameters $(\sigma,M,\lambda)$ for the CG distributions for the different dimensionalities $d$ and sample sizes $n$.}
\label{tab::distributions_parameters_CG}
\end{table}
The plot shows an exponential behavior, with a significant increase starting at $\lambda=10^{-9}$. Based on this trend, reasonable values for $\lambda$ lie between $10^{-6}$ and $10^{-8}$. Once an appropriate range for $\lambda$ is chosen, we vary $M$. In Figure~\ref{fig::comp_test_vs_M_MoG_20D_50K} we show, on the same data, the dependence of the computing time and the average test statistic on $M$ using $\lambda= (10^{-6}, 10^{-7},10^{-8})$.
\begin{figure*}[t!]
    \centering
    \includegraphics[width=0.48\linewidth]{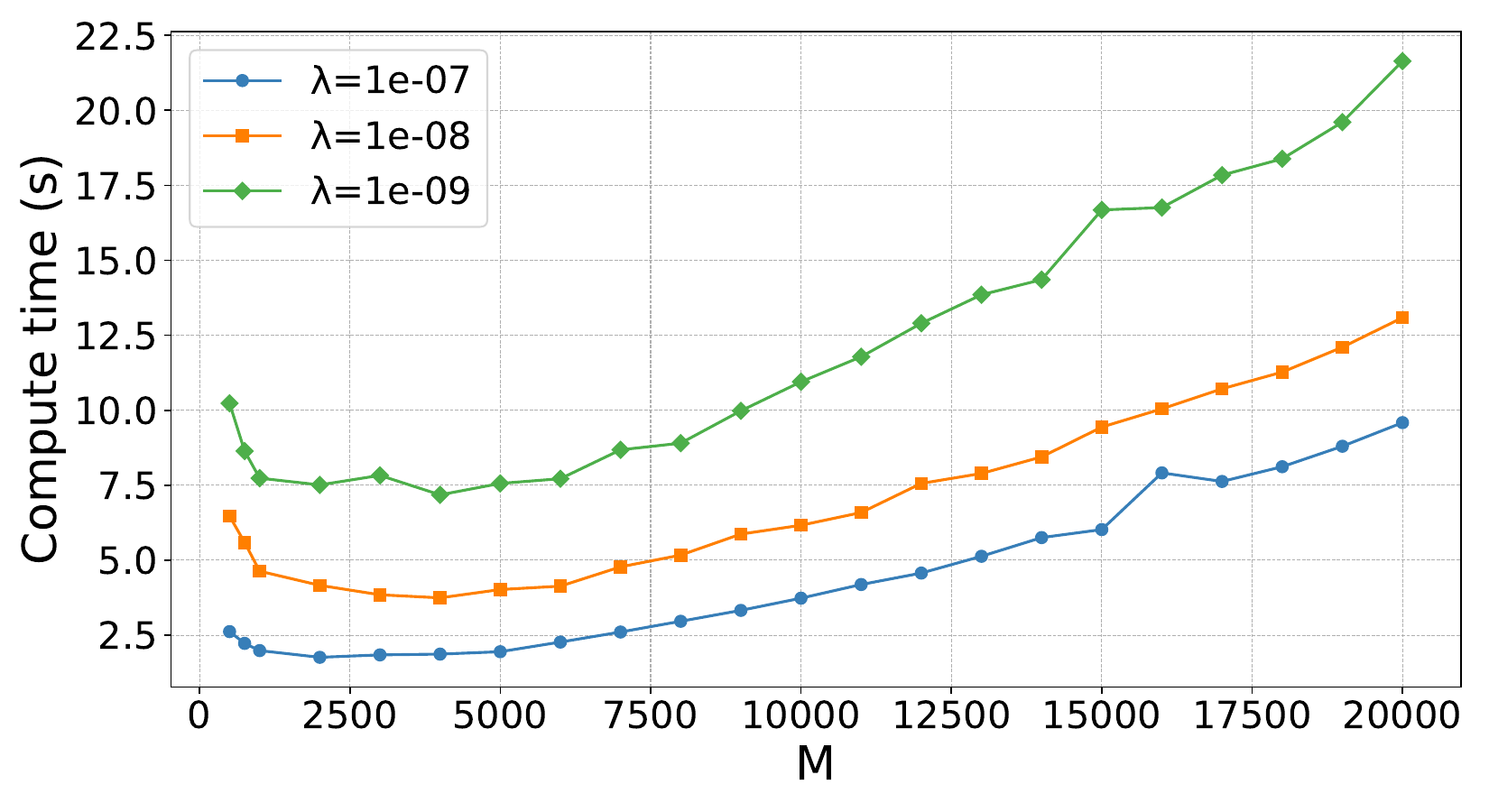}
    \includegraphics[width=0.48\linewidth]{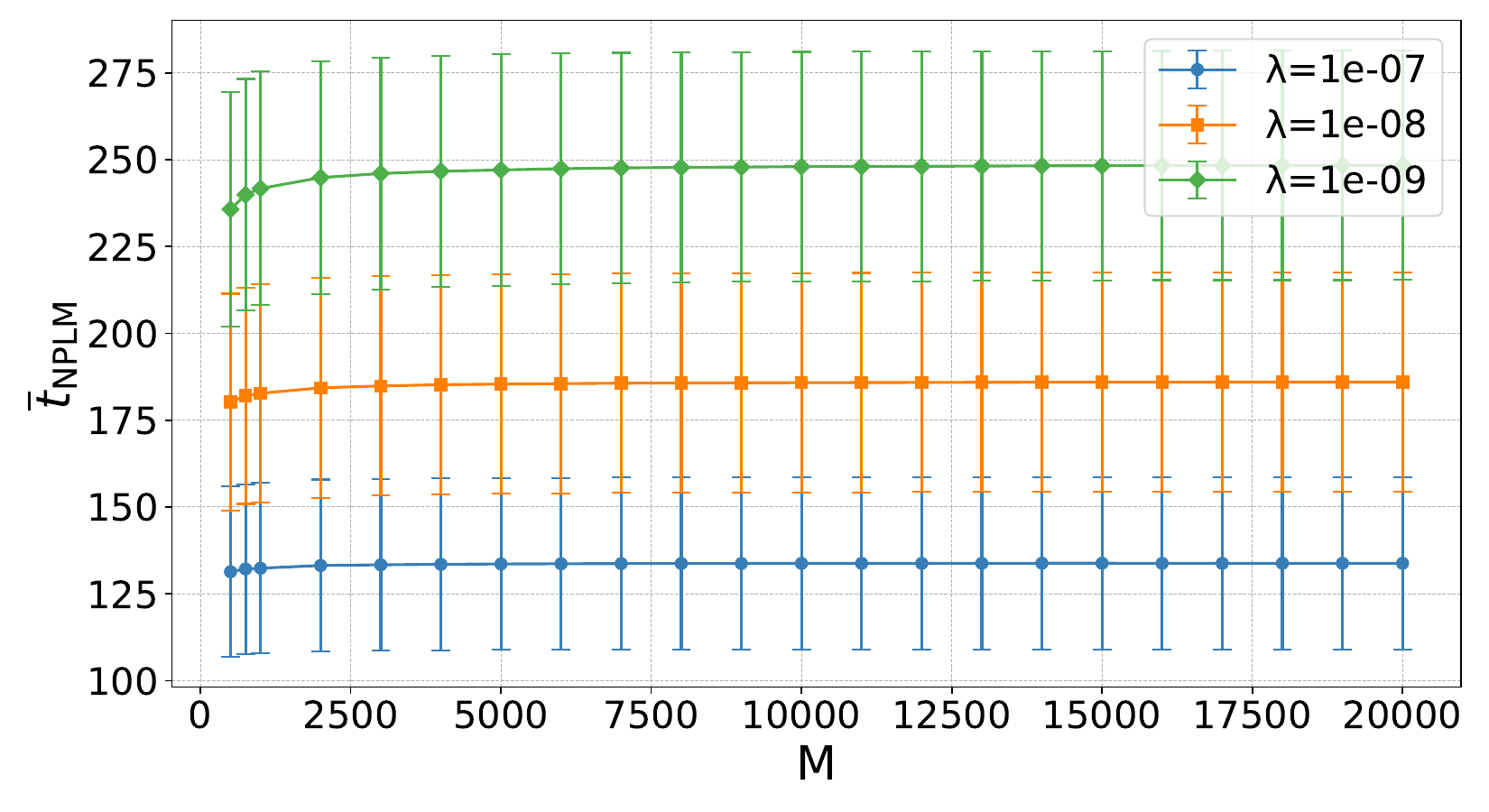}
    \includegraphics[width=0.48\linewidth]{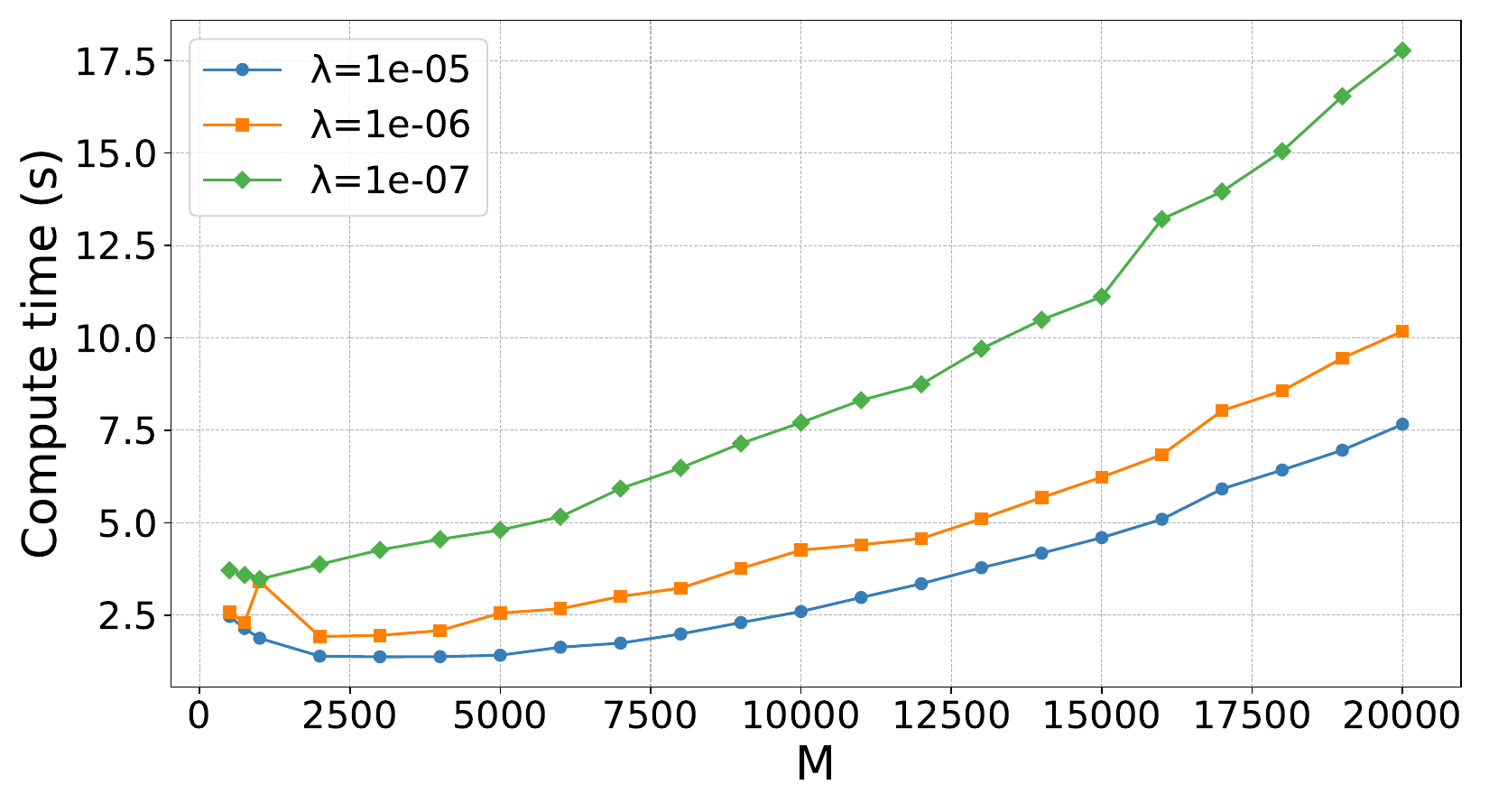}
    \includegraphics[width=0.48\linewidth]{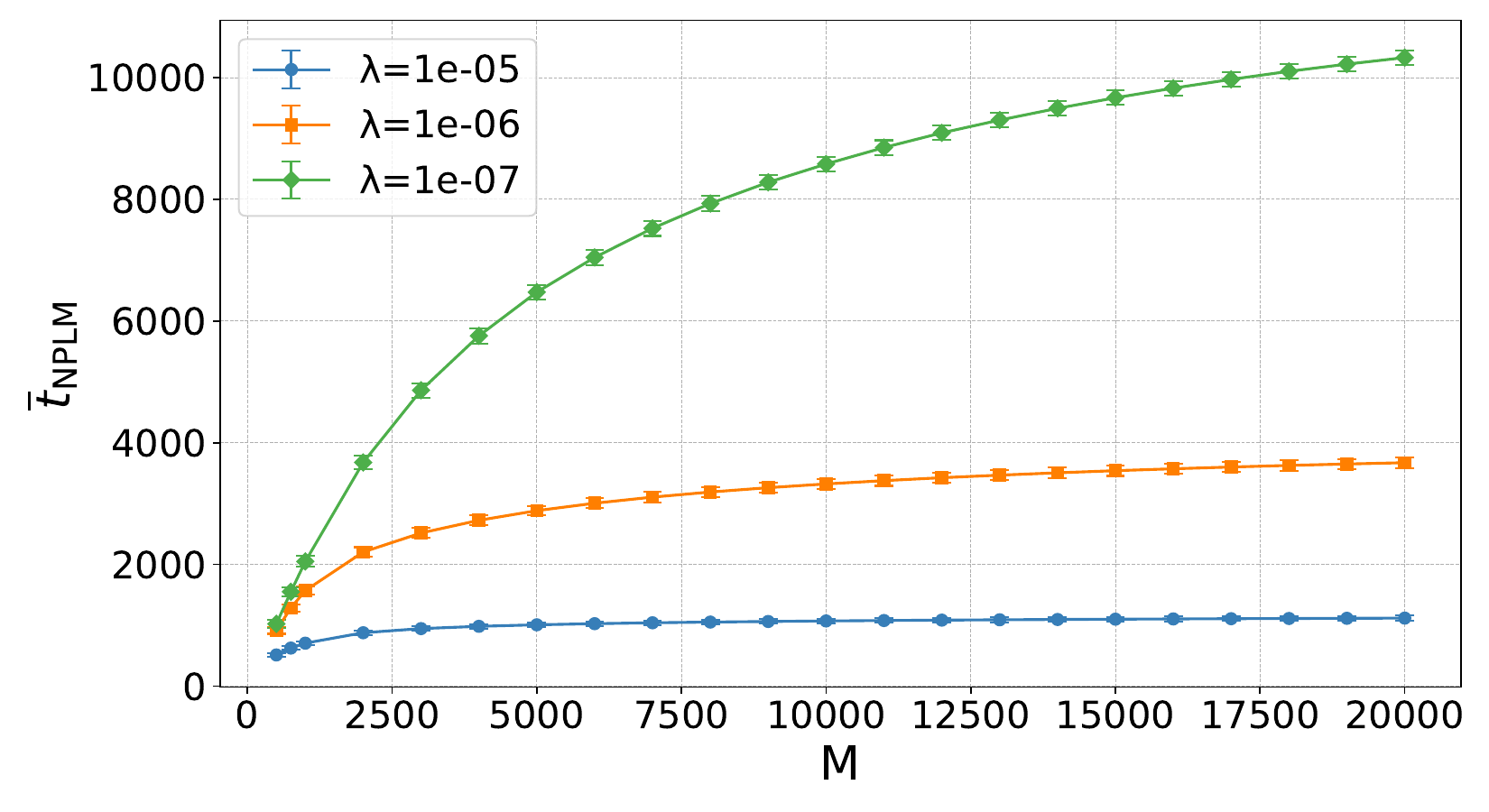}
    \caption{Compute time and mean test statistic as functions of $M$ at varying $\lambda$ for the JetNet dataset. Top row: jet-level features. Bottom row: particle-level features.}
    \label{fig::jet_hyperpar}
\end{figure*}
As expected, computation time increases with $M$ and decreases with $\lambda$. Moreover, the average test statistic stabilizes with a large enough $M$. However, the plateau is reached at higher values as $\lambda$ decreases. In this specific case, we select $(M, \lambda) = (12000, 10^{-7})$, a choice that balances computational efficiency and stability. Note how, by choosing $\lambda=10^{-6}$, the plateau is reached earlier in $M$ but the resulting learning model would be more regularized, hence less expressive.
This procedure has been applied for all the datasets, namely all the other MoG distributions with different dimensionalities and samples sizes, the CG distributions and the particle-level and jet-level datasets from the JetNet dataset.
Ultimately, we find that in most cases the hyperparameters selected for the MoG data were suitable for the CG data, at fixed $d$ and $n$, as shown in Tables~\ref{tab::distributions_parameters_MoG} and~\ref{tab::distributions_parameters_CG}.
The number of Nyström centers decreases with $n$
to maintain a reasonable average computation time. To ensure to be in the plateau of the test statistic as a function of $M$, it is at times needed to increase $\lambda$, given our computational constraints.
In table~\ref{tab::datasets_parameters} we report the hyper-parameters selected for the jet-level and particle-level datasets from the JetNet dataset.
\begin{table}[t!]
\centering
\begin{tabular}{l|c|cc}
	\toprule
	$n$ & \multicolumn{1}{c}{Jet-features} & \multicolumn{2}{c}{Particle-features} \\
    \midrule
	$10K$ & $(4.2,10000,10^{-8})$ & $(18.2,20000,10^{-7})$ \\
	$20K$ & $(4.2,7000,10^{-8})$ & $(18.2,20000,10^{-7})$ \\
	$50K$ & $(4.2,7000,10^{-8})$ &  $(18.2,20000,10^{-7})$ \\
	\bottomrule
\end{tabular}
\caption{{Values of the hyperparameters $(\sigma,M,\lambda)$ for the jet-level and particle-level datasets from the JetNet dataset, for the different sample sizes $n$.}}
\label{tab::datasets_parameters}
\end{table}
For completeness we report in Figure~\ref{fig::jet_hyperpar} the plots which validates the chosen values for the JetNet datasets for the case with $n=20K$.
The whole set of plots for each distribution and dataset can be found in~\cite{NPLM_Parameters_Tuning}.
Overall, we find that model selection is not computationally cheap as it requires multiple evaluations of the test. This needs to be taken into account when deciding which testing method is most suitable for the specific use case. However, datasets characterized by specific dimensionalities and sizes yield models with similar hyperparameters. This suggests that the search for optimal hyperparameters can leverage prior studies, hence mitigating its computational impact.
\begin{figure*}[t!]
    \centering
    \includegraphics[width=0.8\linewidth]{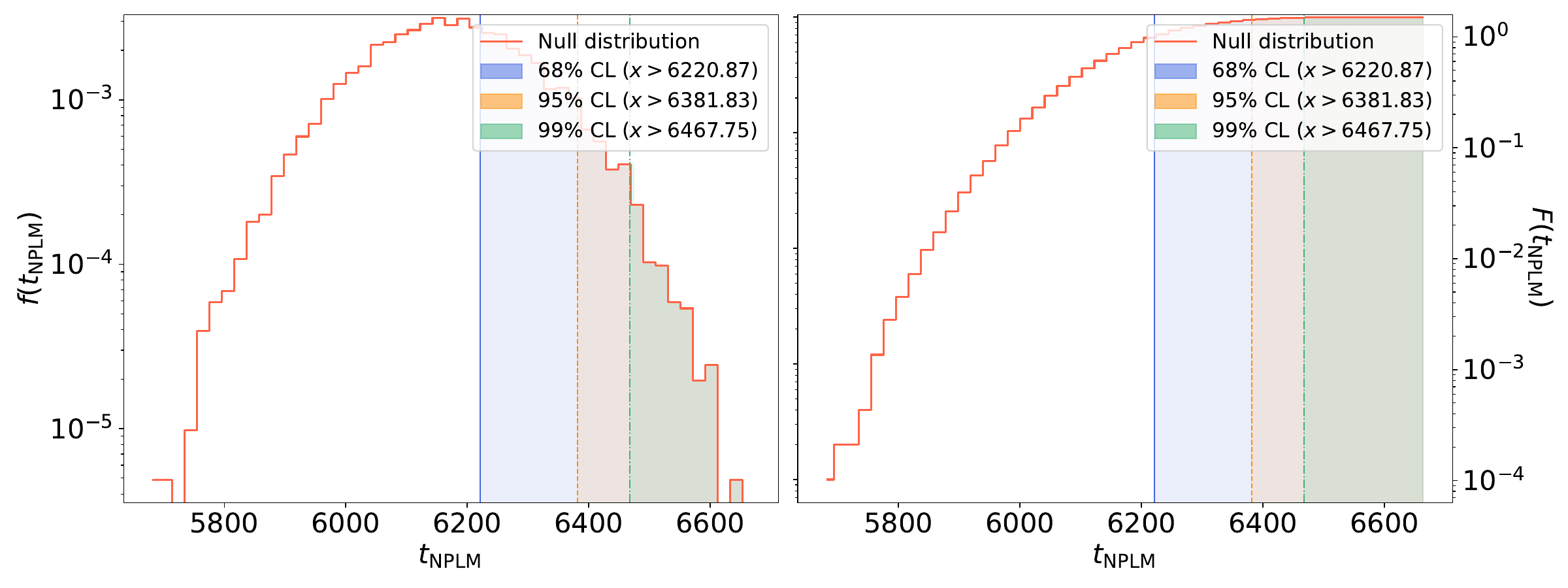}

    \caption{Estimates of the pdf and cdf of the NPLM test statistic under the null hypothesis ($10K$ data points) for the MoG model with $d=20$ and $n=50K$.}
    \label{fig::NPLM_null_MoG_20D_50K}
\end{figure*}
\begin{table*}
  \centering
  \small
  \begin{tabular}{|l|cccccc|}
    \toprule
     Data & $t_{\text{SW}}$ & $t_{\overline{\text{KS}}}$ & $t_{\text{SKS}}$ & $t_{\text{FGD}}$ & $t_{\text{MMD}}$ & $t_{\text{NPLM}}$  \\
    \midrule
    CG ($d=20, n=20k$)  & 0.028   & \textbf{0.011}   & 0.040 & 0.12 & 0.23 & 4.4 \\
    MoG ($d=20, n=50k$) & 0.034   & \textbf{0.016}   & 0.051 & 0.28 & 1.4 & 6.2 \\
    JetNet (jet featues, $n=20k$) & \textbf{0.27}   & 0.32   & 0.87 & 0.50 & 0.95 & 4.5 \\
    JetNet (particle featues, $n=20k$) & \textbf{0.27}   & 0.35   & 0.86 & 1.3 & 0.86 & 6.2 \\
    \bottomrule
  \end{tabular}
  \caption{Mean evaluation time in seconds for a single test (null hypothesis).}
  \label{tab::timing}
\end{table*}

\subsection{Null hypothesis}
Once model selection is performed, we construct the distribution of the test statistic under the null hypothesis $H_{0}$, denoted by $f(t_0)$, by evaluating the test on pairs of samples randomly drawn from the reference distribution. Specifically, we perform 10,000 tests for the Mixture of Gaussians and Correlated Gaussians distributions, and 1,000 for the jet and particle-level features from the JetNet dataset. Once $f(t_0)$ is estimated, we compute the values of the test statistic corresponding to 5\% and 1\% of false positive rates. Figure~\ref{fig::NPLM_null_MoG_20D_50K} shows, as an example, both the pdf $f(t_0)$ and the cdf $F(t_0)$ for the MoG model with dimensionality $d = 20$ and sample size $n = 50K$. The figure also highlights the 32\%, 5\%, and 1\% thresholds.\\

\subsection{Results}
\begin{figure*}[t!]
    \centering    
    \includegraphics[width=0.48\linewidth]{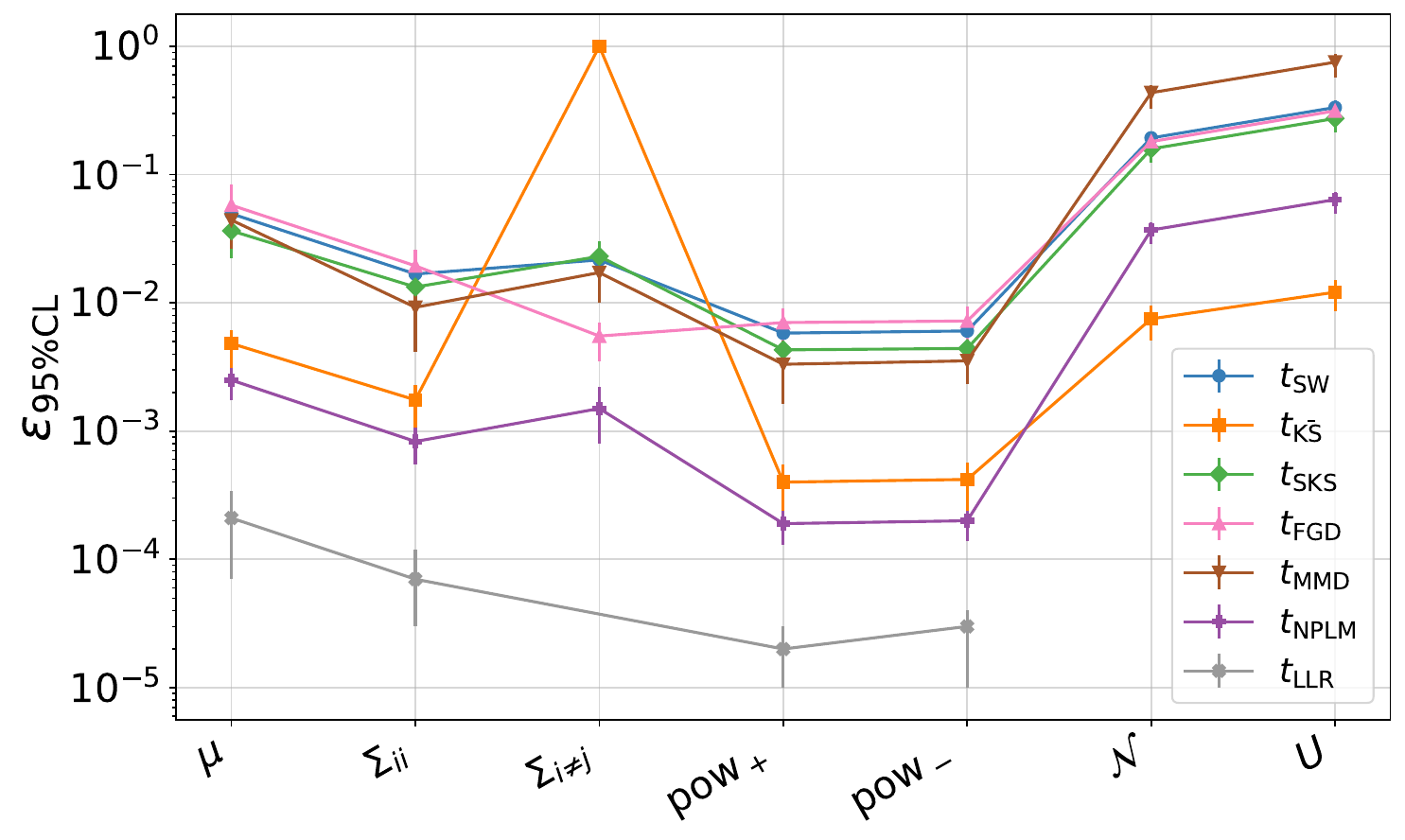}
    \includegraphics[width=0.48\linewidth]{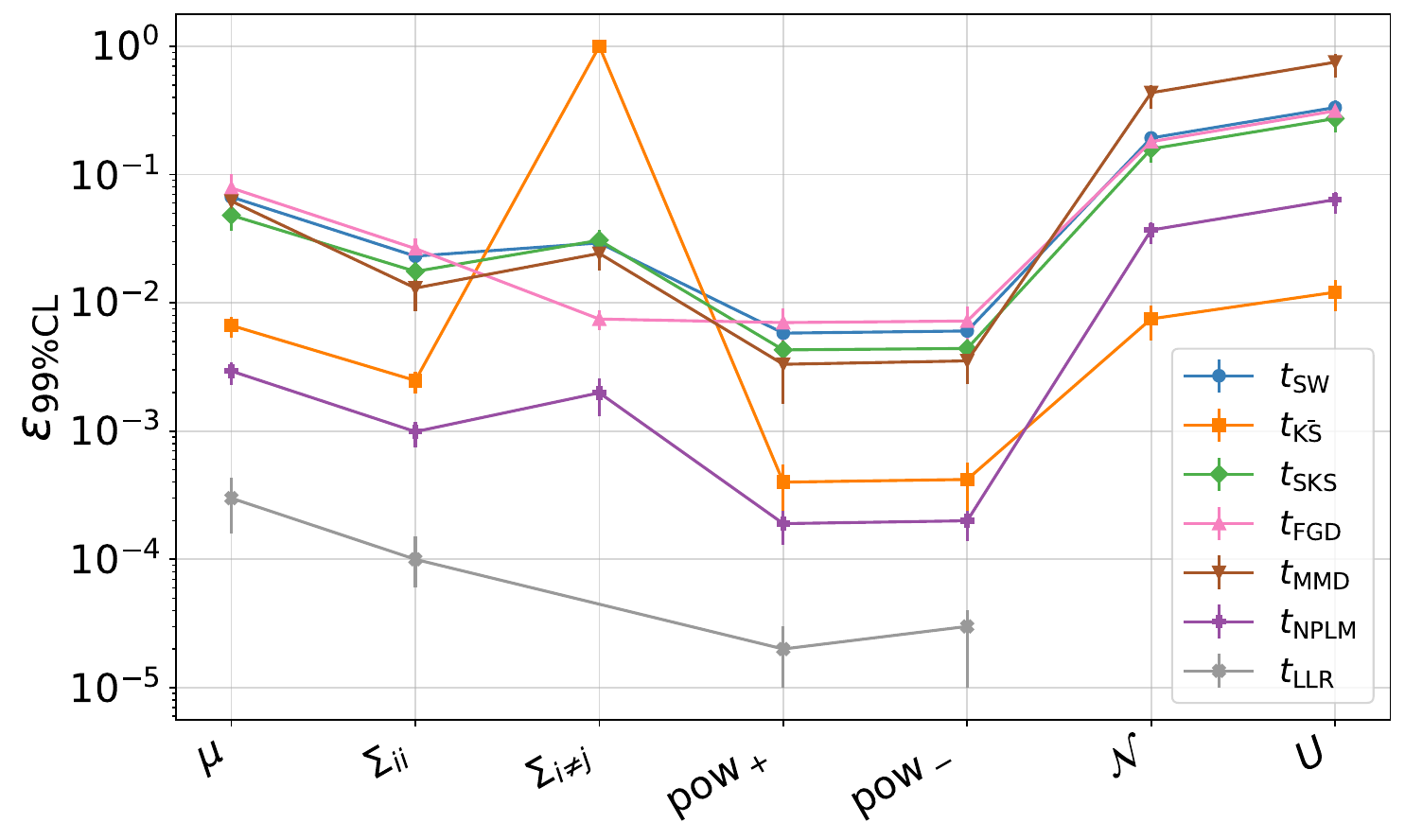}
    \includegraphics[width=0.48\linewidth]{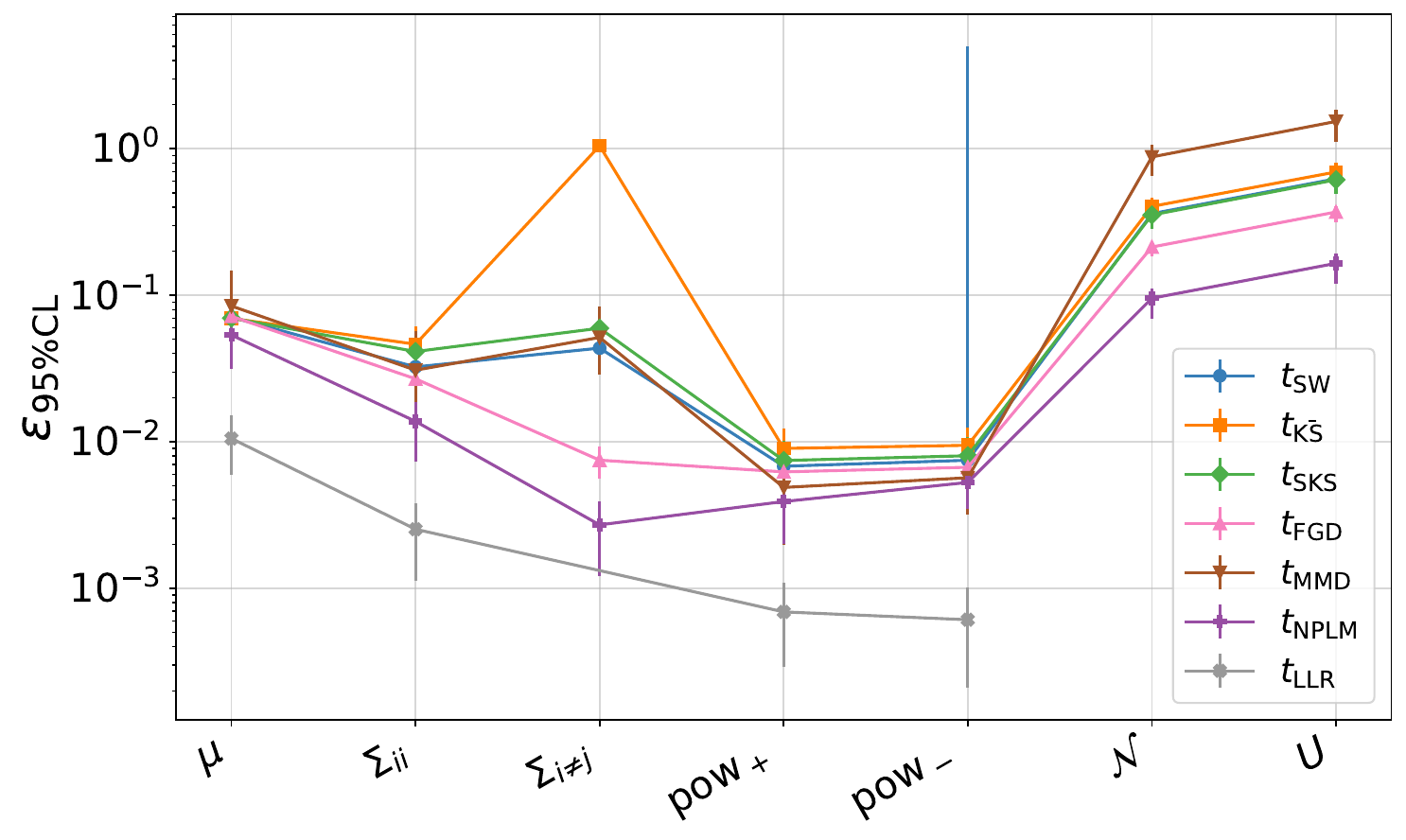}
    \includegraphics[width=0.48\linewidth]{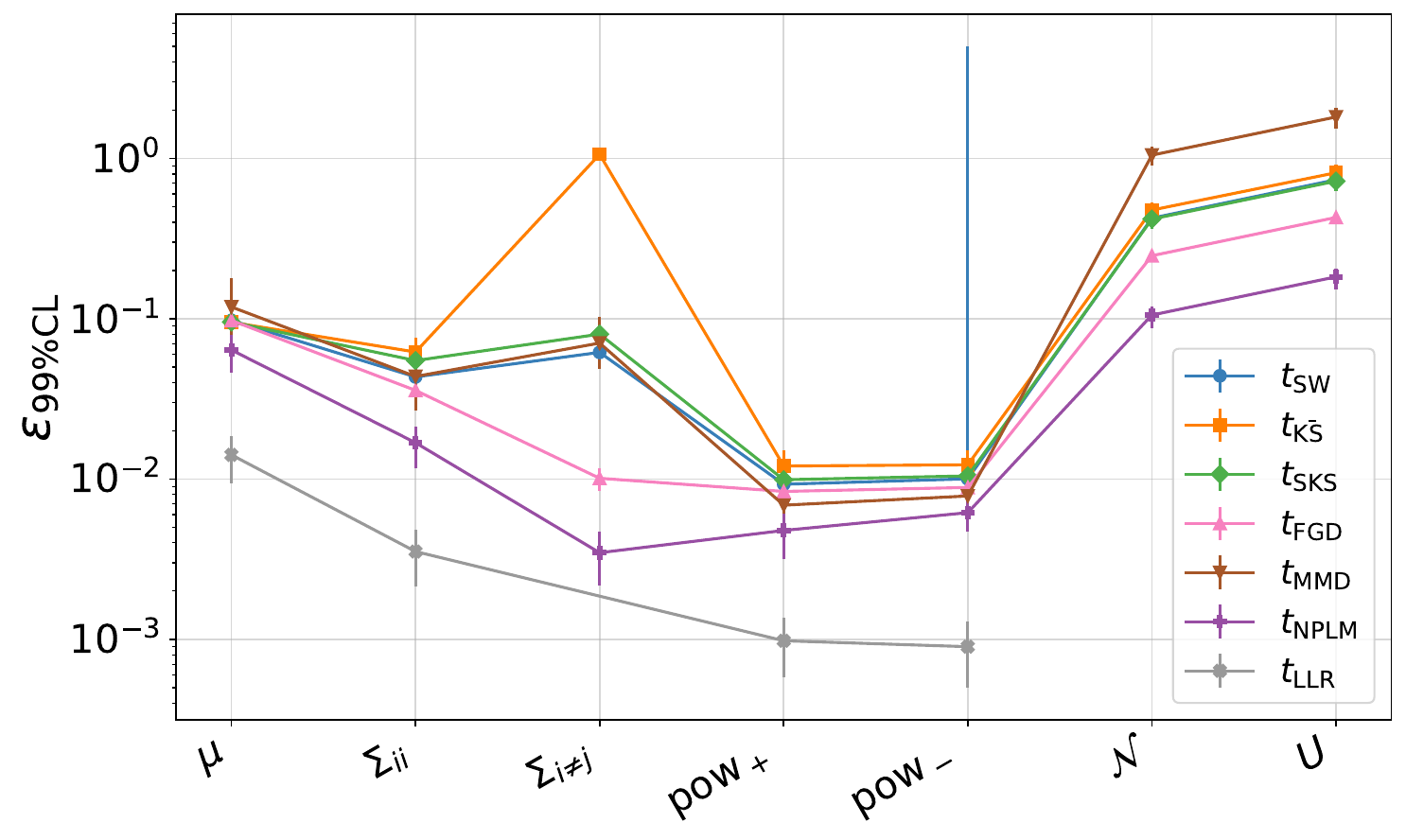}
    \includegraphics[width=0.48\linewidth]{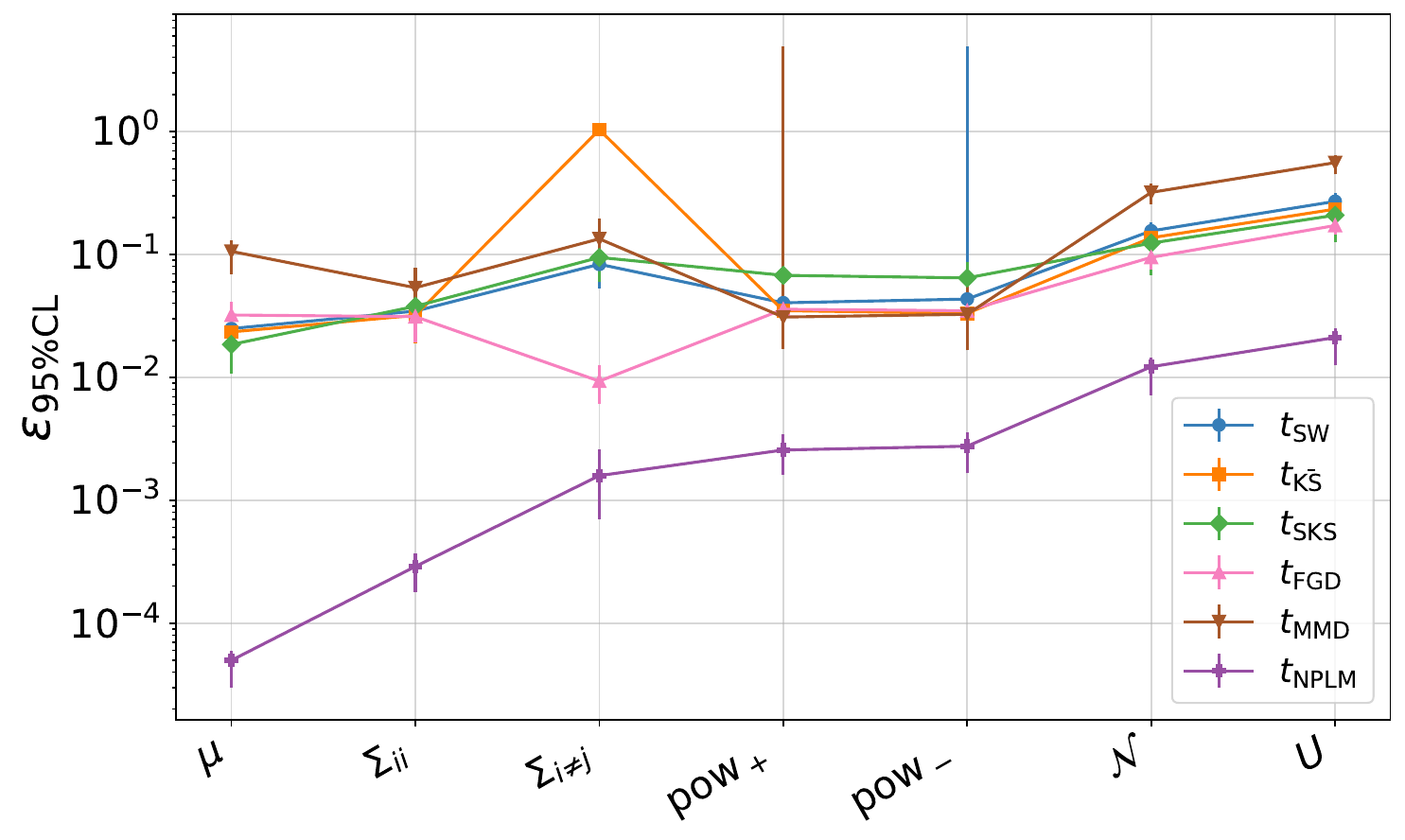}
    \includegraphics[width=0.48\linewidth]{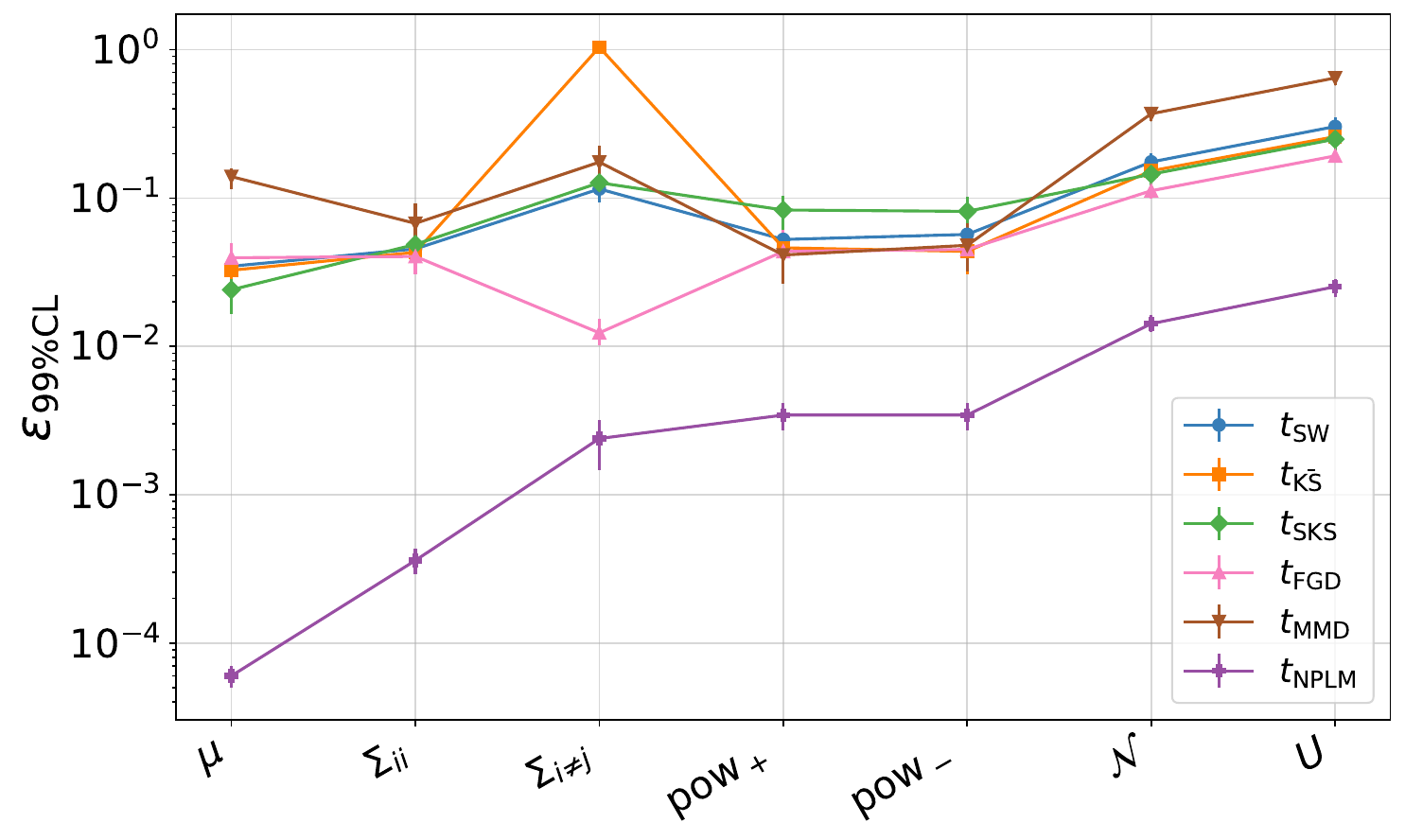}
    \includegraphics[width=0.48\linewidth]{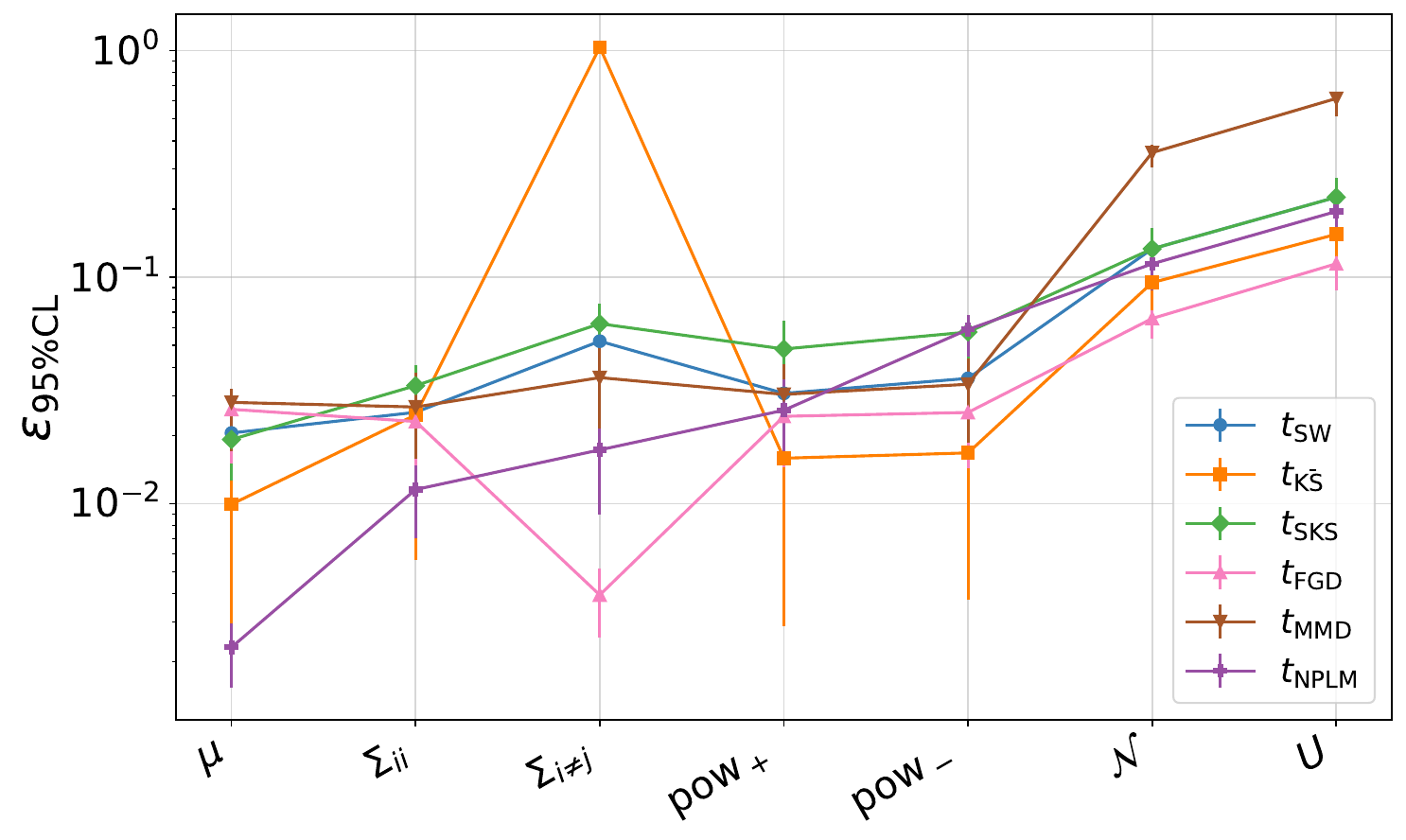}
    \includegraphics[width=0.48\linewidth]{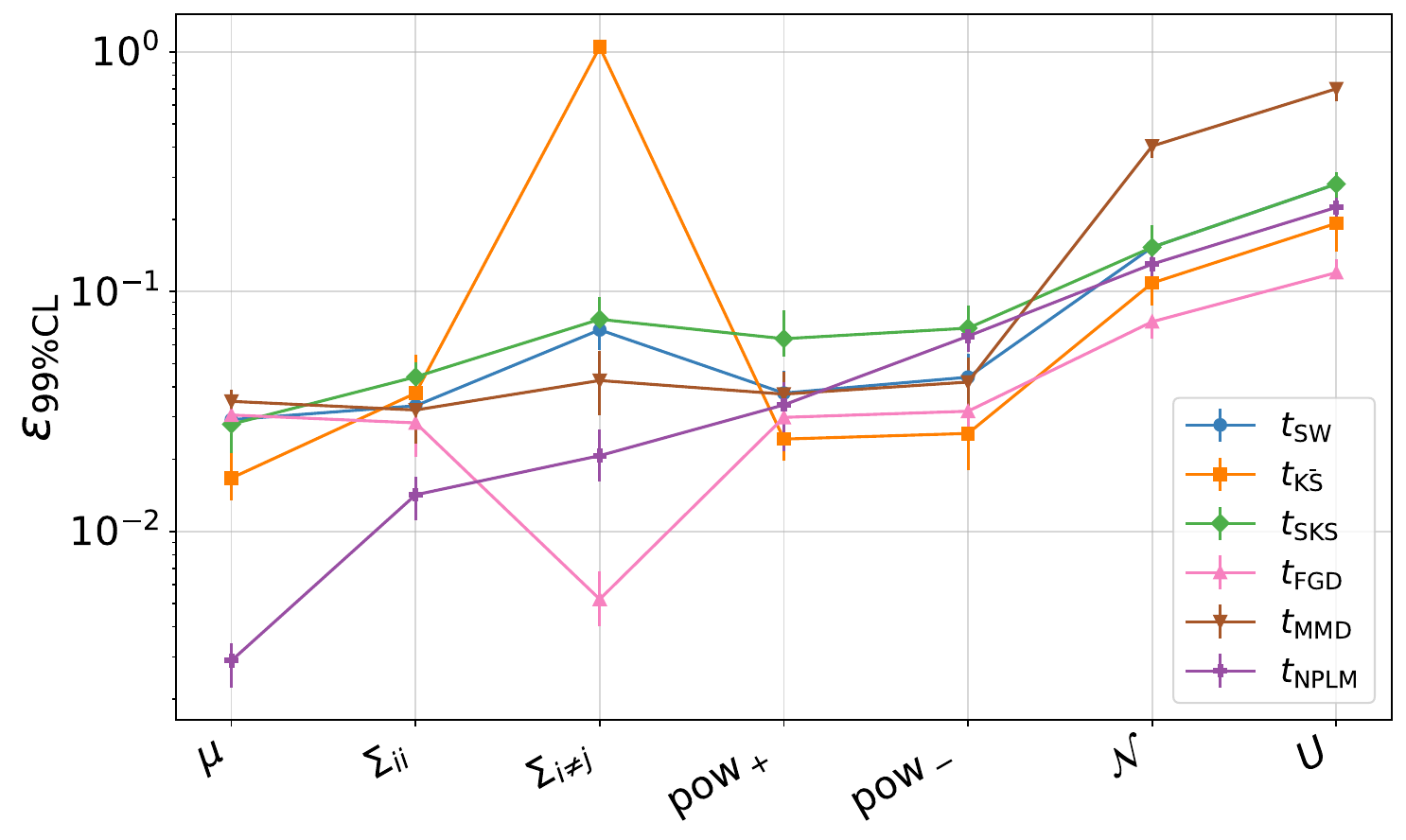}
    \caption{First row: MoG ($d=20$, $n=50k$). Second row: CG ($d=20$, $n=20k$). Third row: JetNet (jet features, $n=20k$. Fourth row: JetNet (particle features, $n=20k$).}
    \label{fig::mogd20n50k}
\end{figure*}

The performance of the NPLM test is compared against the metrics considered in~\cite{Grossi:2024axb}, namely the sliced Wasserstein distance (SW), the Kolmogorov-Smirnov test averaged over marginals ($\overline{KS}$), the sliced Kolmogorov-Smirnov test (SKS), the Fréchet Gaussian Distance (FGD), and the Maximum Mean Discrepancy (MMD). Figure~\ref{fig::mogd20n50k} shows the behavior of the test statistics as functions of the deformations on selected cases for all the datasets. The corresponding numerical values and the complete set of results  can be found in~\ref{app:tables} and in~\cite{NPLM_Gauss_results,NPLM_JetNet_results} respectively.
When available, the exact likelihood ratio test statistic is also reported as a reference, being the most powerful test according to the Neyman-Pearson lemma. We report in Table~\ref{tab::timing} the average time per evaluation for the null hypothesis.

Overall, NPLM is quite robust, with no specific failure cases. It typically ranks as the best or second-best performing metric, within uncertainties. It should be noted that, in this setup, no test is guaranteed to be the most powerful against all alternative hypotheses, and it is always possible for certain tests to outperform others in specific scenarios. In general, it is observed that the performance of the NPLM test is higher at low to intermediate dimensionalities ($d \leq 20$) and improves with increasing sample size. 
This behavior is expected as the performance of kernel methods are known to degrade in high-dimensional settings due to the curse of dimensionality (see however~\cite{Metzger:2025ecl} for recent advances using pretrained deep networks). The advantages of NPLM are also evident when discrepancies involve the correlation structure, this is natural as NPLM is a natively multivariate ML-based approach. This is in contrast to the $\overline{KS}$ test, which is fast and performs well in most cases, but is totally insensitive to the $\Sigma_{i\neq j}$ deformation, as discussed in~\cite{Grossi:2024axb}. All other metrics exhibit similar levels of performance, with FGD being slightly more consistent. However, both MMD and FGD are more computationally expensive to evaluate.
    
\begin{figure}[t!]
    \centering
    {
\includegraphics[width=\linewidth]{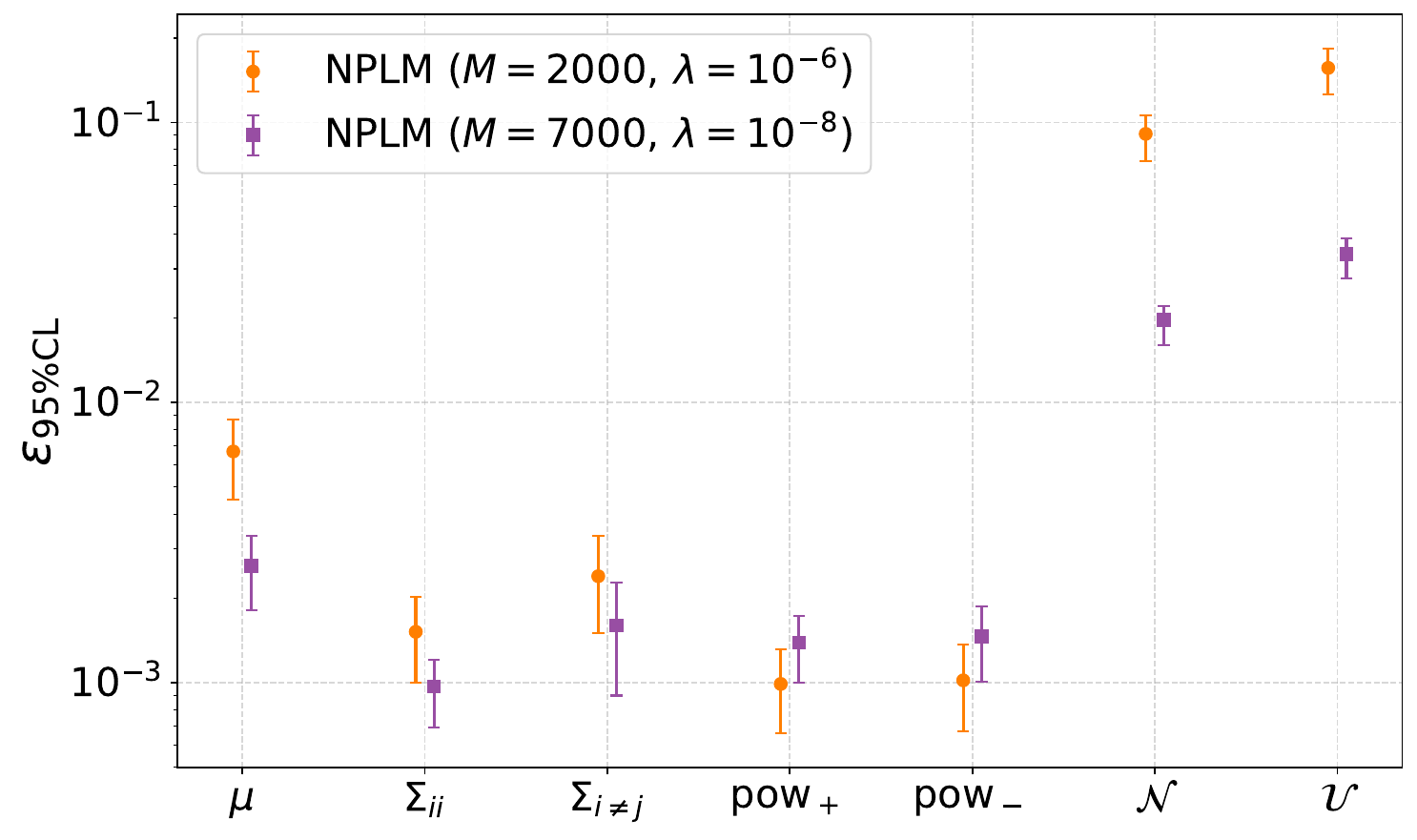}
    }
\caption{Two NPLM models with different hyperparameters on the same data (MoG, $d=5,\,n=20K$). Lower is better.}
    \label{fig::nplm_hyper}
\end{figure}

The relatively high computational cost of NPLM makes it better suited to use cases that do not require ultra-fast evaluations, such as offline data analyses where model complexity can be further increased at the expense of training time. On the other hand, model selection for NPLM can be used to prioritize computational efficiency, as previously shown in~\cite{Grosso:2023ltd}. It is worth mentioning that hyperparameter tuning can have an effect not only on the overall sensitivity of the test but also on the sensitivity to specific alternatives, as shown in Figure~\ref{fig::nplm_hyper}.\footnote{See~\cite{Grosso:2024wjt} for a detailed discussion on this matter and on possible strategies to mitigate this effect.}

\section{Conclusions}
In this work, we tested the NPLM method as a general-purpose two-sample test against a number of alternatives from the literature, following the framework introduced in~\cite{Grossi:2024axb}. We find that NPLM is overall the most robust approach. The method consistently ranks as either the best or second-best performing metric across all tested scenarios, and our study does not highlight specific failure cases. This is relevant, as we expect this advantage to be more pronounced in real-world applications, where potential discrepancies may manifest in multiple simultaneous ways. We find that NPLM is often the best solution for identifying discrepancies in the correlation structure, where it significantly outperforms methods based on marginal comparisons. The performance of the method on the JetNet dataset (jet features) demonstrates its effectiveness on real-world data, achieving sensitivity improvements of up to two orders of magnitude over competing methods. However, in higher-dimensional problems, the differences with other approaches are less pronounced, likely due to the specific implementation based on kernel methods and the impact of the curse of dimensionality (see however~\cite{Metzger:2025ecl} for recent advances using pretrained networks).

This performance, however, comes at the expense of higher computational costs. This cost can be separated into two components: the cost of performing model selection and the cost of executing the test itself. The first involves running a non-negligible number of studies on reference data and is necessary to allow the user to select the optimal working point in the hyperparameter space. This search should be guided by the specific use case. For example, if the goal is to conduct a final offline evaluation of a trained generative model, where fast validation is not critical, then a model that prioritizes performance over efficiency is preferable. However, hyperparameter tuning makes it possible to select a different working point that trades some performance for improved efficiency. As noted in our study, the hyperparameters selected for datasets with similar dimensionalities and sample sizes tend to be close. This suggests that prior studies can be leveraged to reduce the cost of this step. Conversely, if fast evaluation is essential, such as during exploration of the architecture and hyperparameters of the generative model itself, then other approaches such as the $\overline{KS}$ test, the SW test, or the FGD are more appropriate options. Similar considerations apply to the evaluation time of the test, for which NPLM can be between one and three orders of magnitude slower than the fastest approach, depending on the dataset.

Another important point that we only tangentially mentioned in this study is the impact of hyperparameter tuning of ML-powered tests on the sensitivity to specific sources of discrepancies between two generators. While some ML models allow to approximately infer what the impact of model selection will be, modern black box approaches based on neural networks pose significant challenges in this respect (see~\cite{Grosso:2024wjt} for a recent contribution on this topic).


\section*{Acknowledgements}
M.L. acknowledges the financial support of the European Research Council (grant SLING 819789).


\appendix

\section{Deformations}\label{app:deformations}
Various deformations of the reference model are considered, each parametrized by $\epsilon$. The deformations are listed below, using design matrix notation, with $x_{iI}$ and $y_{iI}$ denoting the $I$-th component of the $i$-th point drawn by the reference and deformed model, respectively: 
\begin{enumerate}
    \item $\boldsymbol{\mu}${\bf -deformation} $\longrightarrow$ a shift in the mean: 
    \begin{equation*}
        y_{iI} = x_{iI} + \delta_{\mu\,I}\,,\,\,\,\,  \delta_{\mu\,I} \sim \mathcal{U}_{[-\epsilon,\epsilon]}
    \end{equation*} 
    \item $\boldsymbol{\Sigma_{II}}${\bf -deformation} $\longrightarrow$ a standard deviation increasing: 
    \begin{equation*}
        y_{iI} = \mu_I + c_{\Sigma\,I}(x_{iI}-\mu_I)\,,\,\,\,\,  c_{\Sigma\,I} \sim \mathcal{U}_{[1,1+\epsilon]}
    \end{equation*}
    \item $\boldsymbol{\Sigma_{I\neq J}}${\bf -deformation} $\longrightarrow$  a shrinking of the off-diagonal terms of the correlation matrix:
    \begin{equation*}
        y_{iI} = \sum_j P_{ij}^{(I)}x_{jI}\,,\,\,\,\,  P_{ij}^{(I)}=P_{ij}^{(I)}(\epsilon)\,,\,\,\,\, \text{$P$ permutation matrix}
    \end{equation*}
    \item $\textbf{pow}^{\boldsymbol{+}}${\bf -deformation} $\longrightarrow$  a smearing of each point through a power bigger than one:
    \begin{equation*}
        y_{iI} = \text{sign}(x_{iI})|x_{iI}|^{1+\epsilon}\,,\,\,\,\,  \epsilon \geq 0
    \end{equation*}
    \item $\textbf{pow}^{\boldsymbol{-}}${\bf -deformation} $\longrightarrow$  a smearing of each point through a power smaller than one:
    \begin{equation*}
        y_{iI} = \text{sign}(x_{iI})|x_{iI}|^{1-\epsilon}\,,\,\,\,\, \epsilon \geq 0
    \end{equation*}
    \item $\boldsymbol{\mathcal{N}}${\bf -deformation} $\longrightarrow$  a smearing of each point through a random shift obtained sampling from a normal distribution:
    \begin{equation*}
        y_{iI} = x_{iI} + \delta_{iI}\,,\,\,\,\, \delta_{iI} \sim \mathcal{N}_{0,\epsilon}
    \end{equation*}
    \item $\boldsymbol{\mathcal{U}}${\bf -deformation} $\longrightarrow$  a smearing of each point through a random shift obtained sampling from a uniform distribution:
    \begin{equation*}
        y_{iI} = x_{iI} + \delta_{iI}\,,\,\,\,\, \delta_{iI} \sim \mathcal{U}_{[-\epsilon,\epsilon]}
    \end{equation*}
\end{enumerate}
Each of these deformations is used to define an alternative hypothesis $H_1$, which is tested against $H_{0}$.\\

\section{The NPLM test statistic}\label{app:ext_lr}
In this section we review the derivation of the NPLM test statistic starting from the extended likelihood formalism \cite{barlow1990extended}. Given a model characterized by a pdf $p$ and an expected number of events $n_p$, we can write the likelihood function of a set of observations $\mathcal{Y}=\{y_i\}_{i=1}^m$ as
\begin{equation}
\mathcal{L}_p(\mathcal{Y})=\frac{n_p^m e^{-n_p}}{m!} \prod_{y\in\mathcal{Y}} p(y),
\end{equation}
where the number of observations $m$ is modeled as a Poisson random variable with mean $n_p$. Given a second model with pdf $q$ and expected number of events $n_q$, the log-likelihood ratio test statistic takes the following form
\begin{align}
t(\mathcal{Y})&= -2 \log\frac{\mathcal{L}_p(\mathcal{Y})}{\mathcal{L}_q(\mathcal{Y})}=-2 \log\left[e^{n_q-n_p}\prod_{y\in\mathcal{Y}} \frac{n_q\,q(y)}{n_p\,p(y)}\right]\\\nonumber
&=-2\left[n_q-n_p +\sum_{y\in\mathcal{Y}} \log \frac{n_q\,q(y)}{n_p\,p(y)}\right].
\end{align}
By defining $f(y)= \log \frac{\tilde{q}(y)}{\tilde{p}(y)}$, where $\tilde{p}=(n_q\, q)$ and $\tilde{p}=(n_p\,p)$ are the pdfs normalized to the respective number of events, one obtains
\begin{equation}
t(\mathcal{Y})= -2\left[n_q-n_p -\sum_{y\in\mathcal{Y}} f(y)\right].
\end{equation}
If the density $p$ is not known in closed analytical form but it can be sampled via a generator, we can introduce a reference sample $\mathcal{X}=\{x_i\}_{i=1}^n$ drawn from $p(x)$. Using the definition of $f$, this can be used to estimate $n_q$ as a Monte Carlo sum
\begin{equation}
\begin{split}
n_q & = \int \tilde{q}(x)\,dx = \int \tilde{p} (x)\,e^{f(x)}\,dx\\
&= n_p \int p (x)\,e^{f(x)}\,dx\approx \frac{n_p}{n} \sum_{x\in\mathcal{X}} e^{f(x)}.
\end{split}
\end{equation}
Assuming that $n_p$ is known, the test statistic becomes
\begin{equation}
t(\mathcal{X},\mathcal{Y})= -2\left[\frac{n_p}{n}\sum_{x\in\mathcal{X}} \left(e^{f(x)}-1\right) -\sum_{y\in\mathcal{Y}} f(y)\right].
\end{equation}
If the number of expected events is not a relevant variable, we take $n_p=m$, namely the actual number of data points in the set of observations.

\newpage
\clearpage
\onecolumn
\section{Tables}\label{app:tables}

\begin{table*}[!h]
\centering
\begin{tabular}{l|llr|llr}
	\toprule
	\multicolumn{7}{c}{{\bf CG model with $\mathbf{d=20}$ and $\mathbf{n=m=2\cdot 10^{4}}$}} \\
	\toprule
	\multicolumn{1}{c}{} & \multicolumn{3}{c}{$\mu$-deformation} & \multicolumn{3}{c}{$\Sigma_{ii}$-deformation} \\
	Statistic & $\epsilon_{95\%\mathrm{CL}}$ & $\epsilon_{99\%\mathrm{CL}}$ & $t$ (s) & $\epsilon_{95\%\mathrm{CL}}$ & $\epsilon_{99\%\mathrm{CL}}$ & $t$ (s) \\
	\midrule
	$t_{\mathrm{SW}}$ & $0.07086_{-0.031}^{+0.034}$ & $0.09763_{-0.03}^{+0.033}$ & $496$ & $0.03243_{-0.01}^{+0.0098}$ & $0.04336_{-0.0091}^{+0.0093}$ & $525$ \\
	$t_{\overline{\mathrm{KS}}}$ & $0.06957_{-0.032}^{+0.034}$ & $0.09504_{-0.032}^{+0.032}$ & ${\mathbf{366}}$ & $0.04632_{-0.015}^{+0.015}$ & $0.06199_{-0.014}^{+0.014}$ & ${\mathbf{387}}$ \\
	$t_{\mathrm{SKS}}$ & $0.0699_{-0.031}^{+0.033}$ & $0.09548_{-0.03}^{+0.032}$ & $579$ & $0.04131_{-0.014}^{+0.013}$ & $0.05484_{-0.012}^{+0.012}$ & $620$ \\
	$t_{\mathrm{FGD}}$ & $0.07185_{-0.032}^{+0.041}$ & $0.09756_{-0.03}^{+0.038}$ & $1094$ & $0.0269_{-0.0088}^{+0.01}$ & $0.03574_{-0.0081}^{+0.0091}$ & $1179$ \\
	$t_{\mathrm{MMD}}$ & $0.08449_{-0.049}^{+0.064}$ & $0.11846_{-0.045}^{+0.06}$ & $1574$ & $0.03081_{-0.018}^{+0.026}$ & $0.04364_{-0.017}^{+0.024}$ & $1679$ \\
    $t_{\mathrm{NPLM}}$ & $\mathbf{0.05351_{-0.022}^{+0.017}}$ & $\mathbf{0.06383_{-0.018}^{+0.016}}$ & $23774$ & $\mathbf{0.01378_{-0.0065}^{+0.0048}}$ & $\mathbf{0.01685_{-0.0051}^{+0.0043}}$ & $26822$ \\
	$t_{\mathrm{LLR}}$ & $0.01053_{-0.0046}^{+0.0046}$ & $0.01415_{-0.0047}^{+0.0045}$ & $1306$ & $0.00253_{-0.0014}^{+0.0013}$ & $0.00353_{-0.0014}^{+0.0013}$ & $1504$ \\
	\toprule
	\multicolumn{1}{c}{} & \multicolumn{3}{c}{$\Sigma_{i\neq j}$-deformation} & \multicolumn{3}{c}{$\rm{pow}_{+}$-deformation} \\
	Statistic & $\epsilon_{95\%\mathrm{CL}}$ & $\epsilon_{99\%\mathrm{CL}}$ & $t$ (s) & $\epsilon_{95\%\mathrm{CL}}$ & $\epsilon_{99\%\mathrm{CL}}$ & $t$ (s) \\
	\midrule
	$t_{\mathrm{SW}}$ & $0.04358_{-0.013}^{+0.011}$ & $0.06168_{-0.01}^{+0.0092}$ & ${\mathbf{1024}}$ & $0.00681_{-0.0027}^{+0.0027}$ & $0.00928_{-0.0025}^{+0.0025}$ & $566$ \\
	$t_{\overline{\mathrm{KS}}}$ & $1.04753_{-0.016}^{+0.011}$ & $1.06042_{-0.011}^{+0.017}$ & $1376$ & $0.00901_{-0.0034}^{+0.0033}$ & $0.01206_{-0.0032}^{+0.003}$ & ${\mathbf{422}}$ \\
	$t_{\mathrm{SKS}}$ & $0.05959_{-0.02}^{+0.016}$ & $0.08015_{-0.016}^{+0.015}$ & $1128$ & $0.00745_{-0.0029}^{+0.0027}$ & $0.0099_{-0.0026}^{+0.0025}$ & $632$ \\
	$t_{\mathrm{FGD}}$ & $0.00749_{-0.0019}^{+0.0018}$ & $0.01011_{-0.0017}^{+0.0016}$ & $2468$ & $0.00623_{-0.0025}^{+0.003}$ & $0.00837_{-0.0023}^{+0.0027}$ & $1085$ \\
	$t_{\mathrm{MMD}}$ & $0.05154_{-0.023}^{+0.032}$ & $0.07052_{-0.022}^{+0.032}$ & $2534$ & $0.00488_{-0.0029}^{+0.0042}$ & $0.00687_{-0.0027}^{+0.0039}$ & $1853$ \\
    $t_{\mathrm{NPLM}}$ & $\mathbf{0.00271_{-0.0015}^{+0.0012}}$ & $\mathbf{0.00347_{-0.0013}^{+0.0012}}$ & $36589$ & $\mathbf{0.00391_{-0.0019}^{+0.0015}}$ & $\mathbf{0.00478_{-0.0016}^{+0.0013}}$ & $33561$ \\
	$t_{\mathrm{LLR}}$ & - & - & - & $0.00069_{-0.0004}^{+0.0004}$ & $0.00098_{-0.0004}^{+0.00039}$ & $1628$ \\
	\toprule
	\multicolumn{1}{c}{} & \multicolumn{3}{c}{$\rm{pow}_{-}$-deformation} & \multicolumn{3}{c}{$\mathcal{N}$-deformation} \\
	Statistic & $\epsilon_{95\%\mathrm{CL}}$ & $\epsilon_{99\%\mathrm{CL}}$ & $t$ (s) & $\epsilon_{95\%\mathrm{CL}}$ & $\epsilon_{99\%\mathrm{CL}}$ & $t$ (s) \\
	\midrule
	$t_{\mathrm{SW}}$ & $0.00748_{-0.0026}^{+5}$ & $0.01003_{-0.0024}^{+5}$ & $513$ & $0.36054_{-0.063}^{+0.048}$ & $0.42418_{-0.045}^{+0.041}$ & $472$ \\
	$t_{\overline{\mathrm{KS}}}$ & $0.00946_{-0.0031}^{+0.003}$ & $0.01227_{-0.0028}^{+0.0029}$ & ${\mathbf{418}}$ & $0.40421_{-0.079}^{+0.061}$ & $0.47653_{-0.058}^{+0.053}$ & ${\mathbf{353}}$ \\
	$t_{\mathrm{SKS}}$ & $0.00803_{-0.0029}^{+0.0027}$ & $0.01046_{-0.0026}^{+0.0025}$ & $635$ & $0.35347_{-0.07}^{+0.055}$ & $0.41869_{-0.056}^{+0.042}$ & $519$ \\
	$t_{\mathrm{FGD}}$ & $0.0067_{-0.0023}^{+0.0027}$ & $0.00886_{-0.0021}^{+0.0025}$ & $1090$ & $0.21348_{-0.03}^{+0.022}$ & $0.24723_{-0.019}^{+0.014}$ & $850$ \\
	$t_{\mathrm{MMD}}$ & $0.00568_{-0.0025}^{+0.0037}$ & $0.00784_{-0.0025}^{+0.0036}$ & $1925$ & $0.87964_{-0.23}^{+0.19}$ & $1.04846_{-0.15}^{+0.14}$ & $1413$ \\
    $t_{\mathrm{NPLM}}$ & $\mathbf{0.00528_{-0.0018}^{+0.0015}}$ & $\mathbf{0.00617_{-0.0015}^{+0.0013}}$ & $26885$ & $\mathbf{0.09544_{-0.026}^{+0.016}}$ & $\mathbf{0.10546_{-0.018}^{+0.014}}$ & $21954$ \\
	$t_{\mathrm{LLR}}$ & $0.00061_{-0.0004}^{+0.0004}$ & $0.0009_{-0.0004}^{+0.00039}$ & $1652$ & - & - & - \\
	\toprule
	\multicolumn{1}{c}{} & \multicolumn{3}{c}{$\mathcal{U}$-deformation} & \multicolumn{3}{c}{Timing} \\
	Statistic & $\epsilon_{95\%\mathrm{CL}}$ & $\epsilon_{99\%\mathrm{CL}}$ & $t$ (s) & $t^{\mathrm{null}}$ (s) \\
	\midrule
	$t_{\mathrm{SW}}$ & $0.62405_{-0.1}^{+0.086}$ & $0.73669_{-0.076}^{+0.072}$ & $458$ & $276$ \\
	$t_{\overline{\mathrm{KS}}}$ & $0.69433_{-0.13}^{+0.11}$ & $0.8144_{-0.1}^{+0.099}$ & ${\mathbf{337}}$ & ${\mathbf{108}}$ \\
	$t_{\mathrm{SKS}}$ & $0.61401_{-0.12}^{+0.092}$ & $0.72149_{-0.092}^{+0.078}$ & $505$ & $398$ \\
	$t_{\mathrm{FGD}}$ & $0.37006_{-0.053}^{+0.038}$ & $0.42858_{-0.033}^{+0.025}$ & $812$ & $1150$ \\
	$t_{\mathrm{MMD}}$ & $1.53597_{-0.42}^{+0.32}$ & $1.81389_{-0.27}^{+0.26}$ & $1386$ & $2284$ \\
$t_{\mathrm{NPLM}}$ & $\mathbf{0.16518_{-0.046}^{+0.028}}$ & $\mathbf{0.18251_{-0.031}^{+0.024}}$ & $21414$ & $44027$ \\
	$t_{\mathrm{LLR}}$ & - & - & - & - \\
	\bottomrule
\end{tabular}
\caption{Upper bounds and associated uncertainties on $\epsilon$ at 95\% and 99\% confidence levels, computed for different metrics and deformations. The table also reports the computation times required to estimate these values and to construct the $f(t_0)$ distribution. For each deformation, the best performing metric is indicated in bold.}
\label{tab::results_CG_d=20K_n=20K}
\end{table*}

\begin{table*}
\centering
\begin{tabular}{l|llr|llr}
	\toprule
	\multicolumn{7}{c}{{\bf MoG model with $\mathbf{d=20}$, $\mathbf{q=5}$, and $\mathbf{n=m=5\cdot 10^{4}}$}} \\
	\toprule
	\multicolumn{1}{c}{} & \multicolumn{3}{c}{$\mu$-deformation} & \multicolumn{3}{c}{$\Sigma_{ii}$-deformation} \\
	Statistic & $\epsilon_{95\%\mathrm{CL}}$ & $\epsilon_{99\%\mathrm{CL}}$ & $t$ (s) & $\epsilon_{95\%\mathrm{CL}}$ & $\epsilon_{99\%\mathrm{CL}}$ & $t$ (s) \\
	\midrule
	$t_{\mathrm{SW}}$ & $0.04957_{-0.02}^{+0.018}$ & $0.06694_{-0.017}^{+0.017}$ & $3023$ & $0.01679_{-0.0063}^{+0.005}$ & $0.02315_{-0.005}^{+0.0045}$ & $3197$ \\
	$t_{\overline{\mathrm{KS}}}$ & $0.00482_{-0.0018}^{+0.0013}$ & $0.00667_{-0.0013}^{+0.0011}$ & $2966$ & $0.00175_{-0.00068}^{+0.00052}$ & $0.00248_{-0.00052}^{+0.00042}$ & $3185$ \\
	$t_{\mathrm{SKS}}$ & $0.03647_{-0.014}^{+0.011}$ & $0.04821_{-0.012}^{+0.011}$ & ${\mathbf{2899}}$ & $0.01329_{-0.0043}^{+0.003}$ & $0.01759_{-0.003}^{+0.0025}$ & ${\mathbf{3022}}$ \\
	$t_{\mathrm{FGD}}$ & $0.05778_{-0.027}^{+0.026}$ & $0.0787_{-0.021}^{+0.023}$ & $4047$ & $0.01945_{-0.0081}^{+0.0063}$ & $0.02651_{-0.0056}^{+0.0053}$ & $4507$ \\
	$t_{\mathrm{MMD}}$ & $0.04425_{-0.018}^{+0.019}$ & $0.06215_{-0.015}^{+0.017}$ & $10204$ & $0.00923_{-0.0051}^{+0.0058}$ & $0.01305_{-0.0044}^{+0.0053}$ & $11217$ \\
    $t_{\mathrm{NPLM}}$ & ${\mathbf{0.0025_{-0.00077}^{+0.0006}}}$ & ${\mathbf{0.00294_{-0.00064}^{+0.00051}}}$ & $41843$ & ${\mathbf{0.00083_{-0.00028}^{+0.00023}}}$ & ${\mathbf{0.00099_{-0.00024}^{+0.0002}}}$ & $46776$ \\
	$t_{\mathrm{LLR}}$ & $0.00021_{-0.00014}^{+0.00013}$ & $0.0003_{-0.00014}^{+0.00013}$ & $5911$ & $0.00007_{-0.00004}^{+0.00005}$ & $0.0001_{-0.00004}^{+0.00005}$ & $6304$ \\
	\toprule
	\multicolumn{1}{c}{} & \multicolumn{3}{c}{$\Sigma_{i\neq j}$-deformation} & \multicolumn{3}{c}{$\rm{pow}_{+}$-deformation} \\
	Statistic & $\epsilon_{95\%\mathrm{CL}}$ & $\epsilon_{99\%\mathrm{CL}}$ & $t$ (s) & $\epsilon_{95\%\mathrm{CL}}$ & $\epsilon_{99\%\mathrm{CL}}$ & $t$ (s) \\
	\midrule
	$t_{\mathrm{SW}}$ & $0.02162_{-0.008}^{+0.0056}$ & $0.02935_{-0.0055}^{+0.0045}$ & ${\mathbf{3410}}$ & $0.00581_{-0.0022}^{+0.0017}$ & $0.00798_{-0.0017}^{+0.0015}$ & ${\mathbf{3157}}$ \\
	$t_{\overline{\mathrm{KS}}}$ & $1.00146_{-0.00031}^{+0.00074}$ & $1.00238_{-0.00031}^{+0.00055}$ & $3967$ & $0.0004_{-0.00017}^{+0.00015}$ & $0.00059_{-0.00014}^{+0.00013}$ & $3363$ \\
	$t_{\mathrm{SKS}}$ & $0.02306_{-0.0088}^{+0.0071}$ & $0.03079_{-0.0072}^{+0.0062}$ & $3553$ & $0.0043_{-0.0013}^{+0.0009}$ & $0.00565_{-0.0009}^{+0.00074}$ & $3193$ \\
	$t_{\mathrm{FGD}}$ & $0.00551_{-0.002}^{+0.0015}$ & $0.00748_{-0.0013}^{+0.0013}$ & $6327$ & $0.00702_{-0.0028}^{+0.0021}$ & $0.00965_{-0.0019}^{+0.0016}$ & $4870$ \\
	$t_{\mathrm{MMD}}$ & $0.01723_{-0.0072}^{+0.008}$ & $0.02431_{-0.0064}^{+0.0069}$ & $11450$ & $0.00332_{-0.0017}^{+0.0018}$ & $0.00467_{-0.0014}^{+0.0017}$ & $11801$ \\
    $t_{\mathrm{NPLM}}$ & ${\mathbf{0.0015_{-0.0007}^{+0.0007}}}$ & ${\mathbf{0.00199_{-0.00069}^{+0.0006}}}$ & $112295$ & ${\mathbf{0.00019_{-0.00006}^{+0.00005}}}$ & ${\mathbf{0.00022_{-0.00005}^{+0.00005}}}$ & $52507$ \\
	$t_{\mathrm{LLR}}$ & - & - & - & $0.00002_{-0.00001}^{+0.00001}$ & $0.00002_{-0.00001}^{+0.00001}$ & $6877$ \\
	\toprule
	\multicolumn{1}{c}{} & \multicolumn{3}{c}{$\rm{pow}_{-}$-deformation} & \multicolumn{3}{c}{$\mathcal{N}$-deformation} \\
	Statistic & $\epsilon_{95\%\mathrm{CL}}$ & $\epsilon_{99\%\mathrm{CL}}$ & $t$ (s) & $\epsilon_{95\%\mathrm{CL}}$ & $\epsilon_{99\%\mathrm{CL}}$ & $t$ (s) \\
	\midrule
	$t_{\mathrm{SW}}$ & $0.00604_{-0.0023}^{+0.0017}$ & $0.00825_{-0.0018}^{+0.0016}$ & ${\mathbf{3051}}$ & $0.19318_{-0.039}^{+0.025}$ & $0.22704_{-0.026}^{+0.019}$ & ${\mathbf{2403}}$ \\
	$t_{\overline{\mathrm{KS}}}$ & $0.00042_{-0.00018}^{+0.00015}$ & $0.00061_{-0.00015}^{+0.00013}$ & $3372$ & ${\mathbf{0.00751_{-0.0024}^{+0.002}}}$ & ${\mathbf{0.00993_{-0.002}^{+0.0018}}}$ & $2934$ \\
	$t_{\mathrm{SKS}}$ & $0.00441_{-0.0014}^{+0.00092}$ & $0.00574_{-0.00094}^{+0.00077}$ & $3324$ & $0.15874_{-0.034}^{+0.023}$ & $0.18473_{-0.023}^{+0.019}$ & $2726$ \\
	$t_{\mathrm{FGD}}$ & $0.00722_{-0.0027}^{+0.0021}$ & $0.00987_{-0.0019}^{+0.0016}$ & $4892$ & $0.18095_{-0.038}^{+0.023}$ & $0.21269_{-0.02}^{+0.016}$ & $3756$ \\
	$t_{\mathrm{MMD}}$ & $0.00353_{-0.0015}^{+0.0016}$ & $0.00494_{-0.0012}^{+0.0014}$ & $11418$ & $0.43531_{-0.11}^{+0.066}$ & $0.51609_{-0.054}^{+0.045}$ & $8642$ \\
    $t_{\mathrm{NPLM}}$ & ${\mathbf{0.0002_{-0.00006}^{+0.00004}}}$ & ${\mathbf{0.00023_{-0.00005}^{+0.00004}}}$ & $48084$ & $0.03697_{-0.0081}^{+0.0054}$ & $0.04073_{-0.0056}^{+0.0045}$ & $36180$ \\
	$t_{\mathrm{LLR}}$ & $0.00002_{-0.00001}^{+0.00001}$ & $0.00002_{-0.00001}^{+0.00001}$ & $6991$ & - & - & - \\
	\toprule
	\multicolumn{1}{c}{} & \multicolumn{3}{c}{$\mathcal{U}$-deformation} & \multicolumn{3}{c}{Timing} \\
	Statistic & $\epsilon_{95\%\mathrm{CL}}$ & $\epsilon_{99\%\mathrm{CL}}$ & $t$ (s) & $t^{\mathrm{null}}$ (s) \\
	\midrule
	$t_{\mathrm{SW}}$ & $0.33394_{-0.068}^{+0.044}$ & $0.39248_{-0.044}^{+0.033}$ & ${\mathbf{2354}}$ & $338$ \\
	$t_{\overline{\mathrm{KS}}}$ & ${\mathbf{0.01211_{-0.0035}^{+0.003}}}$ & ${\mathbf{0.01575_{-0.003}^{+0.0027}}}$ & $2835$ & ${\mathbf{155}}$ \\
	$t_{\mathrm{SKS}}$ & $0.27395_{-0.059}^{+0.041}$ & $0.3188_{-0.04}^{+0.033}$ & $2601$ & $509$ \\
	$t_{\mathrm{FGD}}$ & $0.31409_{-0.07}^{+0.04}$ & $0.36919_{-0.036}^{+0.027}$ & $3643$ & $2795$ \\
	$t_{\mathrm{MMD}}$ & $0.75353_{-0.18}^{+0.12}$ & $0.89336_{-0.098}^{+0.078}$ & $7700$ & $13860$ \\
    $t_{\mathrm{NPLM}}$ & $0.06387_{-0.014}^{+0.0096}$ & $0.07083_{-0.01}^{+0.0074}$ & $34879$ & $61789$ \\
	$t_{\mathrm{LLR}}$ & - & - & - & - \\
	\bottomrule
\end{tabular}
\caption{Upper bounds and associated uncertainties on $\epsilon$ at 95\% and 99\% confidence levels, computed for different metrics and deformations. The table also reports the computation times required to estimate these values and to construct the $f(t_0)$ distribution. For each deformation, the best performing metric is indicated in bold.}
\label{tab::results_MoG_d=20K_n=50K}
\end{table*}

\begin{table*}
\centering
\begin{tabular}{l|llr|llr}
	\toprule
	\multicolumn{7}{c}{{\bf JetNet - Jet features with $\mathbf{n=m=2\cdot 10^{4}}$}} \\
	\toprule
	\multicolumn{1}{c}{} & \multicolumn{3}{c}{$\mu$-deformation} & \multicolumn{3}{c}{$\Sigma_{ii}$-deformation} \\
	Statistic & $\epsilon_{95\%\mathrm{CL}}$ & $\epsilon_{99\%\mathrm    {CL}}$ & $t$ (s) & $\epsilon_{95\%\mathrm{CL}}$ & $\epsilon_{99\%\mathrm{CL}}$ & $t$ (s) \\
	\midrule
	$t_{\mathrm{SW}}$ & $0.02498_{-0.0089}^{+0.0068}$ & $0.0347_{-0.007}^{+0.0078}$ & ${\mathbf{1856}}$ & $0.03464_{-0.014}^{+0.011}$ & $0.0454_{-0.012}^{+0.011}$ & ${\mathbf{1983}}$ \\
	$t_{\overline{\mathrm{KS}}}$ & $0.02347_{-0.0084}^{+0.0081}$ & $0.0326_{-0.0065}^{+0.0073}$ & $2379$ & $0.03199_{-0.013}^{+0.012}$ & $0.04302_{-0.012}^{+0.012}$ & $3749$ \\
	$t_{\mathrm{SKS}}$ & $0.01854_{-0.0078}^{+0.0052}$ & $0.02407_{-0.0075}^{+0.0044}$ & $5681$ & $0.03791_{-0.016}^{+0.012}$ & $0.04866_{-0.014}^{+0.012}$ & $6885$ \\
	$t_{\mathrm{FGD}}$ & $0.0322_{-0.013}^{+0.009}$ & $0.03958_{-0.0066}^{+0.01}$ & $4014$ & $0.03122_{-0.012}^{+0.013}$ & $0.04039_{-0.0098}^{+0.012}$ & $3399$ \\
	$t_{\mathrm{MMD}}$ & $0.10604_{-0.037}^{+0.024}$ & $0.13954_{-0.025}^{+0.019}$ & $6850$ & $0.05372_{-0.021}^{+0.025}$ & $0.06748_{-0.021}^{+0.024}$ & $12626$ \\
$t_{\mathrm{NPLM}}$ & $\mathbf{5e-05_{-2e-05}^{+1e-05}}$ & $\mathbf{6e-05_{-1e-05}^{+1e-05}}$ & $34148$ & $\mathbf{0.00029_{-0.00011}^{+8e-05}}$ & $\mathbf{0.00036_{-7e-05}^{+7e-05}}$ & $32391$ \\
	\toprule
	\multicolumn{1}{c}{} & \multicolumn{3}{c}{$\Sigma_{i\neq j}$-deformation} & \multicolumn{3}{c}{$\rm{pow}_{+}$-deformation} \\
Statistic & $\epsilon_{95\%\mathrm{CL}}$ & $\epsilon_{99\%\mathrm{CL}}$ & $t$ (s) & $\epsilon_{95\%\mathrm{CL}}$ & $\epsilon_{99\%\mathrm{CL}}$ & $t$ (s) \\
	\midrule
	$t_{\mathrm{SW}}$ & $0.08331_{-0.03}^{+0.027}$ & $0.11519_{-0.022}^{+0.026}$ & $1581$ & $0.04046_{-0.016}^{+0.015}$ & $0.05256_{-0.016}^{+0.016}$ & ${\mathbf{1560}}$ \\
	$t_{\overline{\mathrm{KS}}}$ & $1.03549_{-0.015}^{+0.0094}$ & $1.04064_{-0.0051}^{+0.015}$ & ${\mathbf{1173}}$ & $0.03495_{-0.014}^{+0.014}$ & $0.04617_{-0.014}^{+0.015}$ & $6217$ \\
	$t_{\mathrm{SKS}}$ & $0.09468_{-0.035}^{+0.035}$ & $0.12683_{-0.024}^{+0.033}$ & $5589$ & $0.06771_{-0.029}^{+0.02}$ & $0.08308_{-0.022}^{+0.02}$ & $9295$ \\
	$t_{\mathrm{FGD}}$ & $0.00933_{-0.0032}^{+0.0032}$ & $0.01233_{-0.0022}^{+0.0031}$ & $6144$ & $0.03586_{-0.016}^{+0.017}$ & $0.04367_{-0.013}^{+0.017}$ & $3613$ \\
	$t_{\mathrm{MMD}}$ & $0.1341_{-0.058}^{+0.062}$ & $0.17463_{-0.046}^{+0.05}$ & $79177$ & $0.03099_{-0.014}^{+4.9}$ & $0.0413_{-0.015}^{+4.9}$ & $13970$ \\
$t_{\mathrm{NPLM}}$ & $\mathbf{0.00159_{-0.00089}^{+0.001}}$ & $\mathbf{0.00239_{-0.00092}^{+0.00081}}$ & $37314$ & $\mathbf{0.00257_{-0.00095}^{+0.00089}}$ & $\mathbf{0.00329_{-0.00069}^{+0.00074}}$ & $35528$ \\
	\toprule
	\multicolumn{1}{c}{} & \multicolumn{3}{c}{$\rm{pow}_{-}$-deformation} & \multicolumn{3}{c}{$\mathcal{N}$-deformation} \\
Statistic & $\epsilon_{95\%\mathrm{CL}}$ & $\epsilon_{99\%\mathrm{CL}}$ & $t$ (s) & $\epsilon_{95\%\mathrm{CL}}$ & $\epsilon_{99\%\mathrm{CL}}$ & $t$ (s) \\
	\midrule
	$t_{\mathrm{SW}}$ & $0.0434_{-0.018}^{+4.9}$ & $0.05684_{-0.016}^{+4.9}$ & ${\mathbf{1381}}$ & $0.15561_{-0.031}^{+0.026}$ & $0.17515_{-0.025}^{+0.026}$ & ${\mathbf{1501}}$ \\
	$t_{\overline{\mathrm{KS}}}$ & $0.03324_{-0.014}^{+0.013}$ & $0.04369_{-0.013}^{+0.015}$ & $7555$ & $0.13682_{-0.027}^{+0.018}$ & $0.15251_{-0.01}^{+0.017}$ & $7585$ \\
	$t_{\mathrm{SKS}}$ & $0.06451_{-0.028}^{+0.022}$ & $0.08132_{-0.021}^{+0.021}$ & $10657$ & $0.12378_{-0.056}^{+0.028}$ & $0.1448_{-0.039}^{+0.028}$ & $11391$ \\
	$t_{\mathrm{FGD}}$ & $0.03489_{-0.015}^{+0.02}$ & $0.04502_{-0.012}^{+0.02}$ & $4270$ & ${\mathbf{0.0948_{-0.02}^{+0.015}}}$ & ${\mathbf{0.11199_{-0.013}^{+0.0096}}}$ & $4014$ \\
	$t_{\mathrm{MMD}}$ & $0.03264_{-0.016}^{+0.024}$ & $0.048_{-0.016}^{+0.02}$ & $15086$ & $0.32021_{-0.066}^{+0.058}$ & $0.37016_{-0.041}^{+0.041}$ & $11399$ \\
$t_{\mathrm{NPLM}}$ & $\mathbf{0.00276_{-0.0011}^{+0.00083}}$ & $\mathbf{0.00344_{-0.00072}^{+0.00071}}$ & $28707$ & $0.01222_{-0.0051}^{+0.0024}$ & $0.01421_{-0.0018}^{+0.0021}$ & $26004$ \\
	\toprule
	\multicolumn{1}{c}{} & \multicolumn{3}{c}{$\mathcal{U}$-deformation} & \multicolumn{3}{c}{Timing} \\
Statistic & $\epsilon_{95\%\mathrm{CL}}$ & $\epsilon_{99\%\mathrm{CL}}$ & $t$ (s) & $t^{\mathrm{null}}$ (s) \\
	\midrule
	$t_{\mathrm{SW}}$ & $0.2705_{-0.059}^{+0.047}$ & $0.30356_{-0.029}^{+0.045}$ & ${\mathbf{1396}}$ & ${\mathbf{271}}$ \\
	$t_{\overline{\mathrm{KS}}}$ & $0.23391_{-0.049}^{+0.024}$ & $0.25939_{-0.024}^{+0.027}$ & $8297$ & $324$ \\
	$t_{\mathrm{SKS}}$ & $0.20877_{-0.082}^{+0.051}$ & $0.24937_{-0.046}^{+0.042}$ & $10459$ & $870$ \\
	$t_{\mathrm{FGD}}$ & $0.17213_{-0.026}^{+0.014}$ & $0.19235_{-0.015}^{+0.016}$ & $2864$ & $497$ \\
	$t_{\mathrm{MMD}}$ & $0.55884_{-0.11}^{+0.088}$ & $0.6454_{-0.07}^{+0.072}$ & $10439$ & $949$ \\
$t_{\mathrm{NPLM}}$ & $\mathbf{0.02112_{-0.0086}^{+0.0039}}$ & $\mathbf{0.02524_{-0.0037}^{+0.003}}$ & $24756$ & $4468$ \\
	\bottomrule
\end{tabular}
\caption{Upper bounds and associated uncertainties on $\epsilon$ at 95\% and 99\% confidence levels, computed for different metrics and deformations. The table also reports the computation times required to estimate these values and to construct the $f(t_0)$ distribution. For each deformation, the best performing metric is indicated in bold.}
\label{tab::results_JN_jet_n=20K}
\end{table*}

\begin{table*}
\centering
\begin{tabular}{l|llr|llr}
	\toprule
	\multicolumn{7}{c}{{\bf Scaled Particle features with $\mathbf{n=m=2\cdot 10^{4}}$}} \\
	\toprule
	\multicolumn{1}{c}{} & \multicolumn{3}{c}{$\mu$-deformation} & \multicolumn{3}{c}{$\Sigma_{ii}$-deformation} \\
	Statistic & $\epsilon_{95\%\mathrm{CL}}$ & $\epsilon_{99\%\mathrm    {CL}}$ & $t$ (s) & $\epsilon_{95\%\mathrm{CL}}$ & $\epsilon_{99\%\mathrm{CL}}$ & $t$ (s) \\
	\midrule
	$t_{\mathrm{SW}}$ & $0.02051_{-0.0073}^{+0.0059}$ & $0.02921_{-0.003}^{+0.0058}$ & ${\mathbf{1482}}$ & $0.02531_{-0.01}^{+0.0089}$ & $0.03339_{-0.0057}^{+0.01}$ & ${\mathbf{1030}}$ \\
	$t_{\overline{\mathrm{KS}}}$ & $0.00993_{-0.0078}^{+0.0052}$ & $0.01668_{-0.0032}^{+0.0052}$ & $1875$ & $0.02465_{-0.019}^{+0.016}$ & $0.03767_{-0.0092}^{+0.017}$ & $2464$ \\
	$t_{\mathrm{SKS}}$ & $0.01919_{-0.0065}^{+0.0052}$ & $0.02798_{-0.0067}^{+0.0013}$ & $3630$ & $0.03319_{-0.013}^{+0.0078}$ & $0.04406_{-0.014}^{+0.0065}$ & $4096$ \\
	$t_{\mathrm{FGD}}$ & $0.02609_{-0.011}^{+0.0054}$ & $0.03056_{-0.0049}^{+0.0057}$ & $4986$ & $0.02305_{-0.0093}^{+0.0077}$ & $0.02833_{-0.0078}^{+0.0068}$ & $6023$ \\
	$t_{\mathrm{MMD}}$ & $0.028_{-0.011}^{+0.0042}$ & $0.03484_{-0.0056}^{+0.0041}$ & $3875$ & $0.02671_{-0.011}^{+0.011}$ & $0.03206_{-0.0088}^{+0.012}$ & $4796$ \\
$t_{\mathrm{NPLM}}$ & $\mathbf{0.00232_{-0.00078}^{+0.00065}}$ & $\mathbf{0.00289_{-0.00066}^{+0.00052}}$ & $34958$ & $\mathbf{0.01154_{-0.0045}^{+0.0032}}$ & $\mathbf{0.0142_{-0.0031}^{+0.0027}}$ & $31575$ \\
	\toprule
	\multicolumn{1}{c}{} & \multicolumn{3}{c}{$\Sigma_{i\neq j}$-deformation} & \multicolumn{3}{c}{$\rm{pow}_{+}$-deformation} \\
Statistic & $\epsilon_{95\%\mathrm{CL}}$ & $\epsilon_{99\%\mathrm{CL}}$ & $t$ (s) & $\epsilon_{95\%\mathrm{CL}}$ & $\epsilon_{99\%\mathrm{CL}}$ & $t$ (s) \\
	\midrule
	$t_{\mathrm{SW}}$ & $0.05216_{-0.021}^{+0.0084}$ & $0.06924_{-0.012}^{+0.01}$ & ${\mathbf{2288}}$ & $0.03068_{-0.014}^{+0.01}$ & $0.03771_{-0.0064}^{+0.009}$ & ${\mathbf{968}}$ \\
	$t_{\overline{\mathrm{KS}}}$ & $1.03785_{-0.024}^{+0.009}$ & $1.04608_{-0.016}^{+0.0082}$ & $2734$ & ${\mathbf{0.01587_{-0.013}^{+0.0085}}}$ & ${\mathbf{0.02425_{-0.0046}^{+0.011}}}$ & $3768$ \\
	$t_{\mathrm{SKS}}$ & $0.06232_{-0.028}^{+0.014}$ & $0.0766_{-0.0079}^{+0.018}$ & $7333$ & $0.04819_{-0.017}^{+0.016}$ & $0.06357_{-0.01}^{+0.02}$ & $5176$ \\
	$t_{\mathrm{FGD}}$ & ${\mathbf{0.00395_{-0.0014}^{+0.0012}}}$ & ${\mathbf{0.00521_{-0.0012}^{+0.0016}}}$ & $10771$ & $0.0243_{-0.0098}^{+0.0079}$ & $0.02987_{-0.0082}^{+0.0072}$ & $4923$ \\
	$t_{\mathrm{MMD}}$ & $0.03602_{-0.015}^{+0.013}$ & $0.04256_{-0.012}^{+0.014}$ & $41440$ & $0.03037_{-0.012}^{+0.011}$ & $0.03734_{-0.011}^{+0.0089}$ & $5925$ \\
$t_{\mathrm{NPLM}}$ & $0.01727_{-0.0083}^{+0.0042}$ & $0.02068_{-0.0045}^{+0.0059}$ & $31422$ & $0.02585_{-0.011}^{+0.0094}$ & $0.03368_{-0.012}^{+0.0081}$ & $29617$ \\
	\toprule
	\multicolumn{1}{c}{} & \multicolumn{3}{c}{$\rm{pow}_{-}$-deformation} & \multicolumn{3}{c}{$\mathcal{N}$-deformation} \\
Statistic & $\epsilon_{95\%\mathrm{CL}}$ & $\epsilon_{99\%\mathrm{CL}}$ & $t$ (s) & $\epsilon_{95\%\mathrm{CL}}$ & $\epsilon_{99\%\mathrm{CL}}$ & $t$ (s) \\
	\midrule
	$t_{\mathrm{SW}}$ & $0.03574_{-0.016}^{+0.0092}$ & $0.04394_{-0.011}^{+0.011}$ & ${\mathbf{973}}$ & $0.13338_{-0.023}^{+0.031}$ & $0.15296_{-0.0054}^{+0.033}$ & ${\mathbf{787}}$ \\
	$t_{\overline{\mathrm{KS}}}$ & ${\mathbf{0.01675_{-0.013}^{+0.0096}}}$ & ${\mathbf{0.0256_{-0.0076}^{+0.011}}}$ & $4374$ & $0.09466_{-0.023}^{+0.019}$ & $0.10855_{-0.021}^{+0.023}$ & $3825$ \\
	$t_{\mathrm{SKS}}$ & $0.05724_{-0.025}^{+0.011}$ & $0.07036_{-0.0072}^{+0.017}$ & $5169$ & $0.13338_{-0.023}^{+0.031}$ & $0.15296_{-0.016}^{+0.033}$ & $4479$ \\
	$t_{\mathrm{FGD}}$ & $0.02527_{-0.011}^{+0.0089}$ & $0.03169_{-0.0076}^{+0.0083}$ & $4913$ & ${\mathbf{0.06571_{-0.012}^{+0.0056}}}$ & ${\mathbf{0.07484_{-0.011}^{+0.003}}}$ & $4537$ \\
	$t_{\mathrm{MMD}}$ & $0.03366_{-0.015}^{+0.01}$ & $0.04192_{-0.01}^{+0.011}$ & $6008$ & $0.35498_{-0.049}^{+0.028}$ & $0.40428_{-0.043}^{+0.016}$ & $4996$ \\
$t_{\mathrm{NPLM}}$ & $0.05875_{-0.014}^{+0.0094}$ & $0.06514_{-0.0092}^{+0.011}$ & $26972$ & $0.11419_{-0.041}^{+0.019}$ & $0.13021_{-0.022}^{+0.017}$ & $25934$ \\
	\toprule
	\multicolumn{1}{c}{} & \multicolumn{3}{c}{$\mathcal{U}$-deformation} & \multicolumn{3}{c}{Timing} \\
Statistic & $\epsilon_{95\%\mathrm{CL}}$ & $\epsilon_{99\%\mathrm{CL}}$ & $t$ (s) & $t^{\mathrm{null}}$ (s) \\
	\midrule
	$t_{\mathrm{SW}}$ & $0.22562_{-0.041}^{+0.048}$ & $0.2808_{-0.036}^{+0.033}$ & ${\mathbf{765}}$ & ${\mathbf{269}}$ \\
	$t_{\overline{\mathrm{KS}}}$ & $0.1547_{-0.043}^{+0.025}$ & $0.19254_{-0.046}^{+0.023}$ & $4034$ & $353$ \\
	$t_{\mathrm{SKS}}$ & $0.22562_{-0.041}^{+0.048}$ & $0.2808_{-0.054}^{+0.033}$ & $4777$ & $857$ \\
	$t_{\mathrm{FGD}}$ & ${\mathbf{0.11465_{-0.027}^{+0.0084}}}$ & ${\mathbf{0.11995_{-0.004}^{+0.016}}}$ & $4174$ & $1336$ \\
	$t_{\mathrm{MMD}}$ & $0.61572_{-0.1}^{+0.041}$ & $0.70123_{-0.081}^{+0.029}$ & $2536$ & $862$ \\
$t_{\mathrm{NPLM}}$ & $0.19517_{-0.068}^{+0.036}$ & $0.22609_{-0.038}^{+0.026}$ & $23624$ & $6158$ \\
	\bottomrule
\end{tabular}
\caption{Upper bounds and associated uncertainties on $\epsilon$ at 95\% and 99\% confidence levels, computed for different metrics and deformations. The table also reports the computation times required to estimate these values and to construct the $f(t_0)$ distribution. For each deformation, the best performing metric is indicated in bold.}
\label{tab::results_JN_par_n=20K}
\end{table*}

\newpage
\clearpage
\twocolumn

\bibliographystyle{elsarticle-harv} 
\bibliography{bibliography}






\end{document}